\documentclass[10pt,twocolumn,letterpaper]{article}

\usepackage[pagenumbers]{cvpr} 

\usepackage[dvipsnames,table]{xcolor}

\definecolor{Graylight}{gray}{0.95}
\definecolor{Grayheavy}{gray}{0.83}
\usepackage{multirow}
\usepackage{adjustbox}
\usepackage{pifont}
\usepackage{arydshln}
\newcommand{\R}[1]{\textcolor[rgb]{1.00,0.00,0.00}{#1}}
\newcommand{\B}[1]{\textcolor[rgb]{0.00,0.00,1.00}{#1}}
\usepackage{multicol}
\usepackage{tabulary}
\usepackage[ruled]{algorithm2e}

\usepackage{graphicx} 
\usepackage{makecell}  
\usepackage{booktabs}    
\usepackage{appendix}
\usepackage{marvosym}

\definecolor{cvprblue}{rgb}{0,0,1}

\usepackage[pagebackref,breaklinks,colorlinks,citecolor=cvprblue]{hyperref}
\def\paperID{***} 
\def\confName{CVPR}
\def\confYear{2024}

\title{\vspace{-0.5em}Multimodal Prompt Perceiver: Empower Adaptiveness, Generalizability and Fidelity for All-in-One Image Restoration\vspace{-0.5em}}

\author{Yuang Ai$^{1,2}$ \quad Huaibo Huang$^{1,2}$\textsuperscript{\Letter} \quad Xiaoqiang Zhou$^{1,3}$ \quad Jiexiang Wang$^{1,3}$ \quad Ran He$^{1,2}$ \\
$^{1}$MAIS \& CRIPAC, Institute of Automation, Chinese Academy of Sciences, Beijing, China \\
$^2$School of Artificial Intelligence, University of Chinese Academy of Sciences, Beijing, China\\
$^{3}$University of Science and Technology of China, Hefei, China\\
\tt\small shallowdream555@gmail.com, \tt\small huaibo.huang@cripac.ia.ac.cn, \\ 
\tt\small\{xq525,jiexiang\}@mail.ustc.edu.cn, rhe@nlpr.ia.ac.cn \\
}

\begin{document}
\twocolumn[{%
  \renewcommand\twocolumn[1][]{#1}%
  \maketitle
  \vspace{-1.15cm}
  \begin{center}
   \centering
   \includegraphics[width=\textwidth]{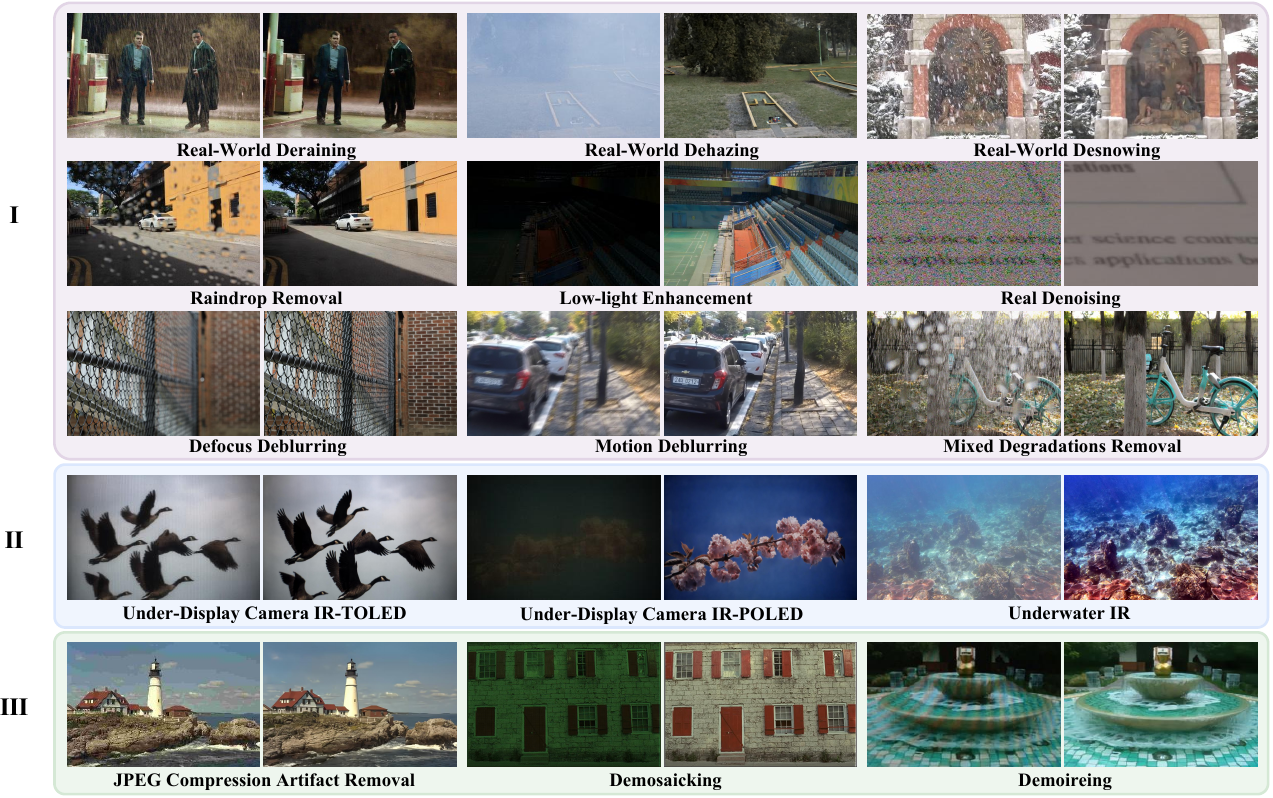}
   \vspace{-0.7cm}
   \captionof{figure}{
   Our \textbf{MPerceiver} excels in image restoration tasks with: \textbf{(\uppercase\expandafter{\romannumeral1}) All-in-one:} Addressing diverse degradations, including challenging mixed ones, through a single pretrained network. \textbf{(\uppercase\expandafter{\romannumeral2}) Zero-shot:} Handling training-unseen degradations effortlessly. \textbf{(\uppercase\expandafter{\romannumeral3}) Few-shot:} Adapting to new tasks with minimal data (about 3\%-5\% of data used by task-specific methods).
   }
   \label{fig:15task}
  \end{center}%
 }]

 \newcommand\blfootnote[1]{%
\begingroup
\renewcommand\thefootnote{}\footnote{#1}%
\addtocounter{footnote}{-1}%
\endgroup
}
\blfootnote{\textsuperscript{\Letter}Corresponding author. \href{https://shallowdream204.github.io/mperceiver/}{Project Page.}}

\begin{abstract}
\vspace{-3mm}
Despite substantial progress, all-in-one image restoration (IR) grapples with persistent challenges in handling intricate real-world degradations. 
This paper introduces \textbf{MPerceiver}: a novel multimodal prompt learning approach that harnesses Stable Diffusion (SD) priors to enhance adaptiveness, generalizability and fidelity for all-in-one image restoration. 
Specifically, we develop a dual-branch module to master two types of SD prompts: textual for holistic representation and visual for multiscale detail representation. 
Both prompts are dynamically adjusted by degradation predictions from the CLIP image encoder, enabling adaptive responses to diverse unknown degradations. 
Moreover, a plug-in detail refinement module improves restoration fidelity via direct encoder-to-decoder information transformation. 
To assess our method, MPerceiver is trained on 9 tasks for all-in-one IR and outperforms state-of-the-art task-specific methods across most tasks. 
Post multitask pre-training, MPerceiver attains a generalized representation in low-level vision, exhibiting remarkable zero-shot and few-shot capabilities in unseen tasks. 
Extensive experiments on 16 IR tasks underscore the superiority of MPerceiver in terms of adaptiveness, generalizability and fidelity.

\end{abstract}    
\section{Introduction}

Image restoration (IR) aims to reconstruct a high-quality (HQ) image from its degraded low-quality (LQ) counterpart. Recent deep learning-based IR approaches excel in addressing single degradation, such as denoising~\cite{zhang2017beyond,zhang2018ffdnet,wang2023lg}, deblurring~\cite{drbnet,stripformer,nrknet}, adverse weather removal~\cite{idt,drsformer,huang2021memory,weatherdiff,li2017aod,snow100k,huang2021selective}, low-light enhancement~\cite{guo2020zero,smg,skf}, \etc. However, these task-specific methods often fall short in real-world scenarios, such as autonomous driving and outdoor surveillance, where images may encounter unknown, dynamic degradations~\cite{mao2017can,zhu2016traffic}.
The concept of all-in-one image restoration has recently gained significant traction, aiming to tackle multiple degradations with a unified model using a single set of pre-trained weights. Leading approaches leverage techniques such as contrastive learning~\cite{airnet,chen2022learning}, task-specific sub-networks~\cite{park2023all,zhu2023learning}, task-specific priors~\cite{idt,valanarasu2022transweather}, and task-agnostic priors~\cite{liu2022tape} to enhance the network's capability across various degradations. Despite their promising performance, the adaptability and generalizability of all-in-one models to real-world scenarios, characterized by intricate and diverse degradations, remain challenging.

Large-scale text-to-image diffusion models, like Stable Diffusion (SD)~\cite{ldm}, succeed in high-quality and diverse image synthesis. This motivates our exploration of leveraging SD for all-in-one image restoration, capitalizing on its HQ image priors to enhance reconstruction quality and generalization across realistic scenarios. However, direct application of SD faces challenges in adaptiveness, generalizability, and fidelity for all-in-one image restoration. SD's proficiency in HQ image synthesis relies on intricately designed prompts, complicating the crafting of suitable prompts for complex, authentic degradations, thereby limiting adaptability and generalization. Furthermore, as a latent diffusion model, SD adopts a VAE architecture with high compression, risking the loss of fine details in restored images and consequently restricting the fidelity of IR~\cite{dai2023emu,zhu2023designing}.

In this paper, we propose MPerceiver, a multimodal prompt learning approach harnessing the generative priors of Stable Diffusion to enhance adaptiveness, generalizability and fidelity of all-in-one image restoration. MPerceiver comprises two modules: a dual-branch module learning textual and visual prompts for diverse degradations, and a detail refinement module (DRM) to boost restoration fidelity. For textual prompt learning, it predicts HQ text embeddings as SD's text condition using CLIP image features from LQ inputs. A cross-modal adapter (CM-Adapter) converts CLIP image embeddings into degradation-aware text vectors, dynamically integrated into HQ textual embeddings based on degradation probabilities estimated by a lightweight predictor. For visual prompts, MPerceiver acquires multiscale detail representations that are crucial for image restoration. An image restoration adapter (IR-Adapter) decomposes VAE image embeddings into multiscale features, dynamically modulated by visual prompts. This dynamic integration in both textual and visual prompt learning enables adaptation to diverse degradations and improves generalization by treating training-unseen degradations as a combination of those in the training set. Additionally, a detail refinement module (DRM) extracts degradation-aware LQ features from the VAE encoder, fused into the decoder through direct encoder-to-decoder information transformation, further enhancing fidelity.

\begin{figure}[!t]
    \centering
    \includegraphics[width=1.0\linewidth]{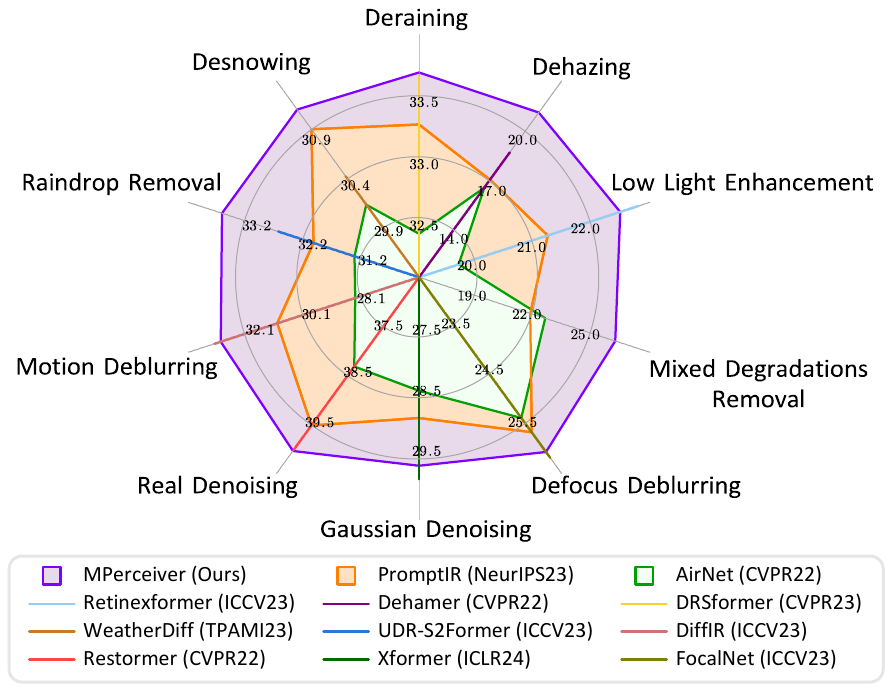}
    \vspace{-7mm}
    \caption{PSNR comparison with state-of-the-art all-in-one and task-specific methods across 10 tasks. Best viewed in color.}
    \label{fig:psnr}
    \vspace{-0.6cm}
\end{figure}

To demonstrate the superiority of MPerceiver as an all-in-one approach, it's trained on 9 IR tasks covering both synthetic and real settings.  As shown in Fig.~\ref{fig:psnr}, our method outperforms all compared all-in-one methods and achieves even better results than state-of-the-art task-specific methods in many tasks. Besides, MPerceiver can even handle challenging mixed degradations that may occur in real-world scenarios (Fig.~\ref{fig:15task} \uppercase\expandafter{\romannumeral1}).
Furthermore, after multitask pre-training, MPerceiver has learned general representations in low-level vision. Comprehensive experiments show that pre-trained MPerceiver exhibits favorable zero-shot and few-shot capabilities in 6 unseen tasks (Fig.~\ref{fig:15task} \uppercase\expandafter{\romannumeral2}, \uppercase\expandafter{\romannumeral3}). 

The main contributions can be summarized as follows:
\begin{itemize}
    \item We propose a novel multimodal prompt learning approach to fully exploit the generative priors of Stable Diffusion for better adaptiveness, generalizability and fidelity of all-in-one image restoration.
    \item We propose a dual-branch module with CM-Adapter and IR-Adapter to learn holistic and multiscale detail representations, respectively. The dynamic integration mechanism for textual and visual prompts enables adaptation to diverse, unknown degradations.
    \item Extensive experiments on 16 IR tasks (all-in-one, zero-shot, few-shot) validate MPerceiver's superiority in achieving high adaptiveness, robust generalizability, and superior fidelity when addressing intricate degradations.
\end{itemize}
\label{sec:intro}

\section{Related Work}
\subsection{Image Restoration}
Image restoration (IR) methods for known degradations~\cite{swinir,mprnet,mou2022deep,chen2021pre,restormer,uformer,nafnet,li2023efficient,zhao2023comprehensive,zhang2022accurate,zhou2023msra,zhou2024ristra,ai2023sosr} have been widely explored, while all-in-one approaches are still in the exploratory stage~\cite{as2020,chen2022learning,liu2022tape,park2023all}. 
TransWeather~\cite{valanarasu2022transweather} designs a transformer-based network with learnable weather type queries to tackle different types of weather. AirNet~\cite{airnet} recovers various degraded images through a contrastive-based degraded encoder. Zhu~\etal~\cite{zhu2023learning} propose a strategy for investigating both weather-general and weather-specific features. IDR~\cite{zhang2023ingredient} employs an ingredients-oriented paradigm to investigate the correlation among various restoration tasks. 
Most of the existing methods are capable of handling a limited range of degradation types and cannot cover complex real-world scenarios.

Since diffusion models have shown a strong capability to generate realistic images~\cite{ddpm,song2020score,beatgan,ramesh2022hierarchical,saharia2022photorealistic}, several diffusion-based methods have been proposed for image restoration~\cite{survey_diff}. These methods can primarily be categorized into zero-shot and supervised learning-based approaches. Zero-shot methods~\cite{DDRM,mainfold,DDNM,chung2022diffusion,GDP,zhu2023denoising,song2022pseudoinverse} leverage pre-trained diffusion models as generative priors, seamlessly incorporating degraded images as conditions into the sampling process. Supervised learning-based methods~\cite{sr3,whang2022deblurring,weatherdiff,saharia2022palette,choi2021ilvr,li2022srdiff} train a conditional diffusion model from scratch. Recently, several approaches~\cite{stablesr,diffbir} have endeavored to employ pre-trained text-to-image diffusion models for blind image super-resolution.  
\subsection{Prompt Learning for Vision Tasks}
Inspired by the success of prompt learning in NLP~\cite{brown2020language,gao2020making,li2021prefix,liu2021p}, the computer vision community has begun to explore its applicability to vision~\cite{jia2022visual,wang2022learning,huang2023diversity,liu2023explicit} and vision-language models~\cite{coop,cocoop,rao2022denseclip,feng2023diverse}. CoOp~\cite{coop} learns a set of task-specific textual prompts to fine-tune CLIP~\cite{clip} for downstream image recognition. CoCoOP~\cite{cocoop} refines the generalizability of CoOp by learning a lightweight neural network to generate image-conditional dynamic prompts. VPT~\cite{jia2022visual} learns a set of visual prompts to finetune transformer-based vision models for downstream recognition tasks. 
Compared to concurrent works that introduce prompt learning into IR~\cite{promptir,prores,daclip,liu2023unifying}, ours is the first to explore multimodal prompt design in low-level vision.

\section{Method}
\begin{figure*}[tb] 
\centering
    \includegraphics[width=1.0\textwidth]{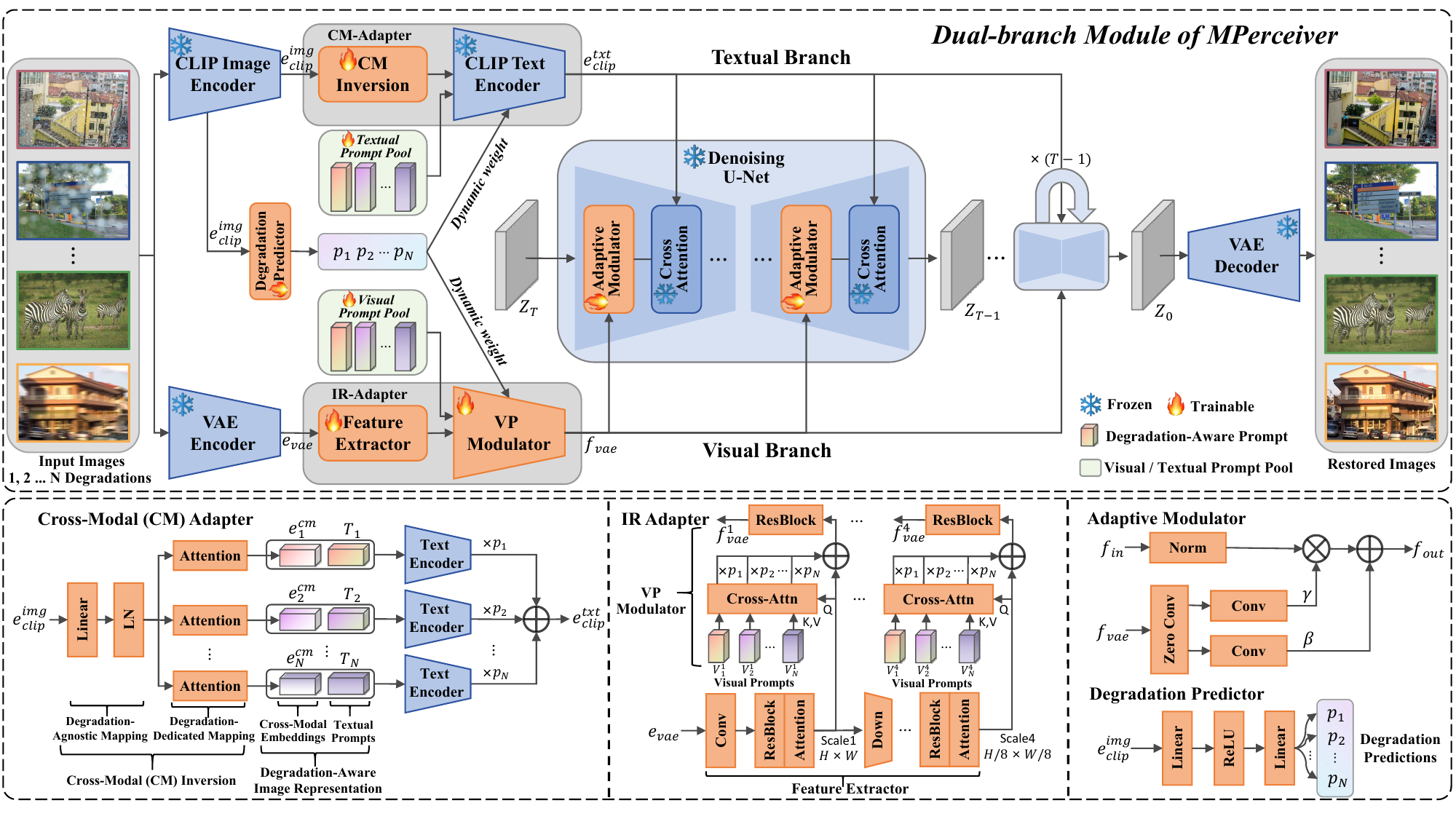}
    \vspace{-7mm}
    \caption{Illustration of MPerceiver's dual-branch module with multimodal prompts. \textit{Textual Branch}: CLIP image embeddings are transformed into text vectors through cross-modal inversion, which are then used alongside textual prompts as holistic representations for SD. \textit{Visual Branch}: IR-Adapter decomposes VAE image embeddings into multi-scale features, which are then dynamically modulated by visual prompts to provide detail guidance for SD adaptively.} \label{fig:arch}
    \vspace{-0.4cm}
\end{figure*}
We propose MPerceiver for all-in-one image restoration in complex real-world scenarios. First, we review the latent diffusion models~\cite{ldm} in Sec.~\ref{sec:preliminary}. To effectively leverage priors in SD, we propose a dual-branch module with the cross-modal adapter (CM-Adapter) and image restoration adapter (IR-Adapter) and encode degradation-dedicated information into multimodal prompts, which is illustrated in Sec.~\ref{sec:dual}. Finally, we introduce a detail refinement module (DRM) to enhance the restoration fidelity in Sec.~\ref{sec:drm}. 

\subsection{Preliminary: Latent Diffusion Models}
\label{sec:preliminary}
Our method is based on Stable Diffusion (SD)~\cite{ldm}, a  text-to-image diffusion model that conducts the diffusion-denoising process in the latent space. SD utilizes a pre-trained VAE to encode images into latent embeddings $z_0$ and then trains the denoising U-Net $\epsilon_\theta$ in the latent space, which can be formulated as 
\begin{equation}
\label{eq:ldm}
\mathcal{L}_{\text{LDM}}=\mathbb{E}_{z_0,c,t,\epsilon}[|| \epsilon-\epsilon_\theta(\sqrt{\bar{\alpha}_t}z_0+\sqrt[]{1-\bar{\alpha }_t}\epsilon ,c,t)||^2_2],
\end{equation}
where $\epsilon \in \mathcal{N}(0,\mathbf{{I}} )$ is the ground truth noise map at time step $t$. $c$ represents the conditional information. $\bar{\alpha}_t$ is the diffusion coefficient in DDPM~\cite{ddpm}.

\subsection{Dual-branch with Multimodal Prompts}
\label{sec:dual}
As shown in Fig.~\ref{fig:arch}, the proposed MPerceiver adopts a dual-branch (\ie, textual branch and visual branch) module with textual and visual prompts in their corresponding branch. Motivated by CLIP's powerful representation capability for images~\cite{distill2021multimodal,hernandez2021natural,materzynska2022disentangling}, we utilize the pre-trained image encoder $\mathcal{E}_{clip}^{img}$ to extract features rich in degradation-aware information. The degraded features $e_{clip}^{img}$ will be fed into a lightweight trainable degradation predictor to provide predictions $P \in \mathbb{R}^N$ ($N$ denotes the number of degradations), which will serve as dynamic weights to adjust the integration process of multimodal prompts. The degradation predictor is optimized through focal loss~\cite{focalloss}.

\noindent \textbf{Textual branch with CM-Adapter.}
To effectively leverage the powerful text-to-image generation capability of SD, we aim to obtain the text description of desired HQ images. As shown in Fig.~\ref{fig:arch}, we propose a cross-modal adapter (CM-Adapter) with the cross-modal inversion mechanism to transform CLIP LQ image embeddings $e_{clip}^{img}$ to desired HQ text embeddings $e_{clip}^{txt}$.

Specifically, LQ image embeddings $e_{clip}^{img}$ will first go through a small network for degradation-agnostic mapping. Then we employ a series of parallel self-attention~\cite{attention} layers as the degradation-dedicated mapping  to obtain one set of degradation-dedicated cross-modal embeddings (i.e., text vectors) $e^{cm}=\left \{ e^{cm}_1,\cdots, e^{cm}_N\right  \}$, where $e^{cm}_i$ corresponds to a specific type of degradation. The whole process of cross-modal inversion can be formulated as:
\begin{equation}
    e^{cm}_i=\mathrm{Attn}_i(\mathrm{LN}(\mathrm{FC}(e^{img}_{clip}))),i\in\{1,\cdots ,N\},
\end{equation}
where $\mathrm{Attn}_i$ is the self-attention layer for the $i$-th degradation, and $N$ is the number of degradations. 

Furthermore, we establish a textual prompt (TP) pool $T = \{T_1,\cdots,T_N\}\in \mathbb{R}^{N\times L\times C_{clip}}$ to encapsulate degradation-dedicated information, where $L$ represents the number of tokens, and $C_{clip}$ is the embedding dimension of CLIP. Serving as learnable parameters, $T$ collaborates with cross-modal embeddings $e^{cm}$ to constitute a comprehensive set of SD text prompts, functioning as representations aware of image degradations. Given degradation probabilities $P=\left \{ p_1,\cdots, p_N\right  \}\in \mathbb{R}^N$ estimated by the degradation predictor, we dynamically integrate the set of text prompts to obtain the High-Quality (HQ) text embeddings:
\begin{equation}
    \label{eq:cm-adapter}
    e_{clip}^{txt}= \sum_{i=1}^{N}p_i\mathcal{E}_{clip}^{txt}(\{e^{cm}_i,T_i\}).
\end{equation} 
where $\mathcal{E}_{clip}^{txt}$ denotes the CLIP text encoder. Finally, we integrate   $e_{clip}^{txt}$ into SD through a frozen cross-attention layer to provide a holistic representation.

\noindent \textbf{Visual branch with IR-Adapter.}
The textual branch can provide a holistic representation for SD, but it lacks detailed information that is crucial for restoration fidelity.  A visual branch is introduced to extract multi-scale detail representations and complement with the textual branch. As shown in Fig.~\ref{fig:arch}, we first project degraded images into latent embeddings $e_{vae}\in \mathbb{R}^{H\times W\times 4}$ through the VAE encoder of SD. 
Then we propose an image restoration adapter (IR-Adapter) to acquire multi-scale detail features as guidance for SD. We utilize a feature extractor to decompose $e_{vae}$ into multi-scale features $f_{vae}=\{f_{vae}^1,f_{vae}^2,f_{vae}^3,f_{vae}^4\}$. In each scale, we employ a residual block (RB) and a self-attention layer to extract features. Similar to the textual prompt pool, we construct 4 visual prompt pools $V=\{V^k,|k\in\{1,2,3,4\}\}$ for each scale, where $V^k \in\mathbb{R}^{N\times M \times C_k}$, $M$ is a hyper-parameter specifying the capacity of visual prompt (VP) pools and $C_k$ is the channel dimension of the $k$-th scale.  As shown in Fig.~\ref{fig:arch}, we propose a visual prompt (VP) modulator to dynamically integrate the degradation-aware information provided by visual prompts into multi-scale features,  formulated as:
\begin{equation}
    \label{eq:ir-adapter}
    f^k_{vae}=\mathrm{RB}_k(f^k_{vae}+\sum_{i=1}^Np_i\mathrm{MHCA}_k(f^k_{vae},V^k_i,V^k_i)),
\end{equation}
where $\mathrm{MHCA}_k(q,k,v)$ is the multi-head cross-attention layer of the $k$-th scale. Then the degradation-aware multi-scale features will modulate the features in the U-Net with the operation of AdaIN~\cite{huang2017arbitrary},  formulated as:
\begin{equation}
    f_{out} =\gamma (f_{vae})\odot \mathrm{Norm}(f_{in})+\beta(f_{vae}),
\end{equation}
where $\odot$ denotes the element-wise multiplication, $f_{in}$ is the original feature in the U-Net, $f_{out}$ is the feature after modulation, $\gamma(\cdot)$ and $\beta(\cdot)$ are implemented by two convolutional layers. The dual-branch module is optimized directly using the latent diffusion loss in Eq.~(\ref{eq:ldm}). 

Note that the dynamic integration mechanisms in Eq.~(\ref{eq:cm-adapter}) and Eq.~(\ref{eq:ir-adapter}) significantly augment the adaptiveness and generalizability of MPerceiver when confronting diverse degradations. In the case of degradations present during training, the model can discern their types and select corresponding textual and visual prompts for Stable Diffusion (SD). For those unseen during training (especially for mixed ones that often occur in real-world scenarios), it treats them as a probabilistic combination of known training degradations.

\subsection{Detail Refinement Module}
\label{sec:drm}
\begin{figure}[t]
\centering
\hspace{-1.7mm}
\begin{subfigure}[b]{0.32\linewidth}
\centering
   \includegraphics[width=1\linewidth]{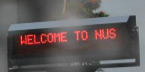}
\end{subfigure}
\begin{subfigure}[b]{0.32\linewidth}
\centering
   \includegraphics[width=1\linewidth]{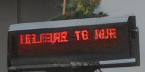}
\end{subfigure}
\begin{subfigure}[b]{0.32\linewidth}
\centering
   \includegraphics[width=1\linewidth]{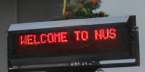}
\end{subfigure}
\begin{subfigure}[b]{0.32\linewidth}
\centering
   \includegraphics[width=1\linewidth]{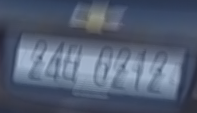}
   \caption{Input}
\end{subfigure}
\begin{subfigure}[b]{0.32\linewidth}
\centering
   \includegraphics[width=1\linewidth]{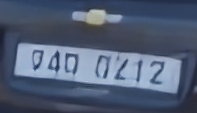}
   \caption{w/o DRM}
\end{subfigure}
\begin{subfigure}[b]{0.32\linewidth}
\centering
   \includegraphics[width=1\linewidth]{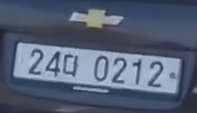}
   \caption{w/ DRM}
\end{subfigure}
\vspace{-3mm}
\caption{Effect of the DRM on raindrop removal (top row from~\cite{raindrop}) and motion deblurring (bottom row from~\cite{gopro}). The proposed DRM significantly improves the fidelity of the results.}
\vspace{-4.5mm}
\label{fig:drm_ab}
\end{figure}

While the proposed dual-branch module empowers Stable Diffusion (SD) with a robust ability to identify and eliminate degradations in images, the generated images may exhibit a tendency to lose details of small objects. This effect is attributed to the high compression rate of the autoencoder employed by SD, as discussed in previous works~\cite{dai2023emu,zhu2023designing}. For instance, in Fig.~\ref{fig:drm_ab} (b), although the degradations in the images have been largely addressed, noticeable artifacts become apparent, particularly affecting small text.

\begin{figure}[t]
    \centering
    \includegraphics[width=1.0\linewidth]{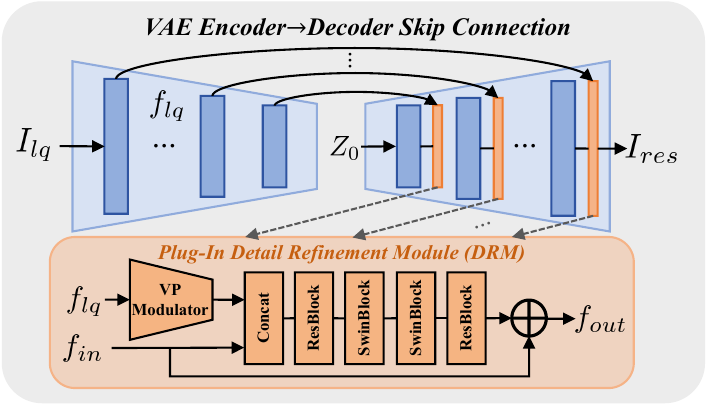}
    \vspace{-7mm}
    \caption{Illustration of the detail refinement module (DRM). For simplicity, visual prompts and degradation predictions are omitted as input to the visual prompt (VP) modulator. Apart from the DRM which is trainable, the other modules are all frozen.}
    \vspace{-4mm}
    \label{fig:drm}
\end{figure}

To address this concern, we introduce a Detail Refinement Module (DRM) aimed at providing supplementary information to assist the SD VAE decoder in the image reconstruction process. As depicted in Fig.~\ref{fig:drm}, DRM functions as a plug-in module, enabling direct encoder-to-decoder information transformation through a skip connection. Following modulation by the visual prompt (VP) modulator, Low-Quality (LQ) features $f_{lq}$ are concatenated with the original decoder features $f_{in}$. Subsequently, a sequence of ResBlocks~\cite{nafnet} and SwinBlocks~\cite{swinir} is employed to extract auxiliary features, enhancing detail reconstruction before the residual connection.
The training of  DRM involves a combination of reconstruction (L1) loss, color loss~\cite{wang2019underexposed}, perceptual loss~\cite{johnson2016perceptual}, and adversarial loss~\cite{goodfellow2014generative}. 

\section{Experiments}
\begin{table*}[t]
  \caption{\textbf{[All-in-one]} Quantitative comparison with state-of-the-art task-specific methods and all-in-one methods on 9 tasks. General IR models trained under the all-in-one setting are marked with symbol $(\cdot)_A$. Best and second best performance are in \R{red} and \B{blue} colors, respectively. When using the self-ensemble strategy, the model is marked with ``+''.}
  \vspace{-3mm}
 \label{tab:compare_9task}
  \centering
  \renewcommand\arraystretch{0.88}
  \fontsize{6.6pt}{\baselineskip}\selectfont
  \setlength\tabcolsep{1.2pt}
  \begin{tabular}{cccccccccc}
  \toprule
  \hline
   \multicolumn{1}{c|}{\multirow{+2}*{Type}} &\multicolumn{1}{c|}{\multirow{+2}*{Method}} & \multicolumn{2}{c|}{\textbf{\textit{Deraining}} (Rain1400)} & 
  \multicolumn{1}{c|}{\multirow{+2}*{Method}} & \multicolumn{2}{c|}{\textbf{\textit{Dehazing}} (Average)} & 
  \multicolumn{1}{c|}{\multirow{+2}*{Method}} & \multicolumn{2}{c}{\textbf{\textit{Desnowing}} (Snow100K-L)} \\ \cline{3-4} \cline{6-7} \cline{9-10}
   \multicolumn{1}{c|}{\multirow{-2}*{Type}} & \multicolumn{1}{c|}{\multirow{-2}*{Method}} &  ~~~PSNR / SSIM $\uparrow$ & \multicolumn{1}{c|}{~~~~~FID / LPIPS $\downarrow$}& 
  \multicolumn{1}{c|}{\multirow{-2}*{Method}} &  ~~~PSNR / SSIM $\uparrow$ & \multicolumn{1}{c|}{~~~~~FID / LPIPS $\downarrow$}& 
  \multicolumn{1}{c|}{\multirow{-2}*{Method}} & ~~~PSNR / SSIM $\uparrow$ & ~~~~~FID / LPIPS $\downarrow$               \\ \hline
  \multicolumn{1}{c|}{\multirow{4}{*}{\begin{tabular}[c]{@{}c@{}}Task \\ Specific\end{tabular}}} & \multicolumn{1}{c|}{Uformer~\cite{uformer}} 
  & 32.84 / 0.931          
  & \multicolumn{1}{c|}{{23.31 / 0.061}} 
  & \multicolumn{1}{c|}{AECRNet~\cite{aecrnet}} 
  & 17.84 / 0.546        
  & \multicolumn{1}{c|}{225.8 / 0.526}               
  & \multicolumn{1}{c|}{DesnowNet~\cite{snow100k}}    
  &  27.17/ 0.898
  & - / -
  \\
   \multicolumn{1}{c|}{} & \multicolumn{1}{c|}{Restormer~\cite{restormer}}
  & \B{33.68} / \B{0.939}          
  & \multicolumn{1}{c|}{20.33 / 0.050}          
  & \multicolumn{1}{c|}{SGID~\cite{sgid}}          
  & 14.36 / 0.562        
  & \multicolumn{1}{c|}{342.9 / 0.580}               
  & \multicolumn{1}{c|}{DDMSNet~\cite{ddmsnet}} 
  & 28.85 / 0.877
  & ~3.24 / 0.096
  \\
  \multicolumn{1}{c|}{}& \multicolumn{1}{c|}{DRSformer~\cite{drsformer}}
  & 33.66 / \B{0.939}          
  & \multicolumn{1}{c|}{20.06 / 0.050}          
  & \multicolumn{1}{c|}{DeHamer~\cite{dehamer}}            
  & 18.64 / 0.622                        
  & \multicolumn{1}{c|}{241.6 / 0.488}               
  & \multicolumn{1}{c|}{DRT~\cite{drt}}     
  & 29.56 / 0.892
  & ~8.15 / 0.135
  \\
  \multicolumn{1}{c|}{}& \multicolumn{1}{c|}{UDR-S$^{2}$~\cite{udrs2former}}
  & 33.08 / 0.930          
  & \multicolumn{1}{c|}{19.89 / 0.053}          
  & \multicolumn{1}{c|}{MB-Taylor~\cite{taylorformer}}         
  & 17.94 / 0.602                
  & \multicolumn{1}{c|}{250.8 / 0.499}              
  & \multicolumn{1}{c|}{WeatherDiff~\cite{weatherdiff}}
  & 30.43 / 0.915
  & ~2.81 / 0.100
  \\
  \hline
  \multicolumn{1}{c|}{\multirow{7}{*}{\begin{tabular}[c]{@{}c@{}}All \\ in \\One \end{tabular}}} & \multicolumn{1}{c|}{AirNet~\cite{airnet}}
  & 32.36 / 0.928                    
  & \multicolumn{1}{c|}{22.38 / 0.058}          
  & \multicolumn{1}{c|}{AirNet~\cite{airnet}}           
  & 16.48 / 0.589                         
  & \multicolumn{1}{c|}{219.9 / 0.479}              
  & \multicolumn{1}{c|}{AirNet~\cite{airnet}}
  & 30.14 / 0.907
  & ~3.92 / 0.105
  \\
    \multicolumn{1}{c|}{}& \multicolumn{1}{c|}{PromptIR~\cite{promptir}} 
  & 33.26 / 0.935              
  & \multicolumn{1}{c|}{22.59 / 0.058} 
  & \multicolumn{1}{c|}{PromptIR~\cite{promptir}}  
  & 16.97 / 0.595   
  & \multicolumn{1}{c|}{231.9 / 0.471}            
  & \multicolumn{1}{c|}{PromptIR~\cite{promptir}}
  & 30.91 / 0.913
  & ~3.79 / 0.100
  \\
      \multicolumn{1}{c|}{}& \multicolumn{1}{c|}{DA-CLIP~\cite{daclip}} 
  & 29.67 / 0.851              
  & \multicolumn{1}{c|}{35.01 / 0.116} 
  & \multicolumn{1}{c|}{DA-CLIP~\cite{daclip}}  
  & 15.01 / 0.544   
  & \multicolumn{1}{c|}{224.6 / 0.468}            
  & \multicolumn{1}{c|}{DA-CLIP~\cite{daclip}}
  & 28.31 / 0.862
  & ~3.11 / 0.098
  \\
\multicolumn{1}{c|}{}& \multicolumn{1}{c|}{Restormer$_A$~\cite{restormer}}
  & 33.09 / 0.933         
  & \multicolumn{1}{c|}{24.14 / 0.061}          
  & \multicolumn{1}{c|}{Restormer$_A$~\cite{restormer}}         
  & 15.86 / 0.584             
  & \multicolumn{1}{c|}{221.4 / 0.477}               
  & \multicolumn{1}{c|}{Restormer$_A$~\cite{restormer}}
  & 30.98 / 0.914
  & ~4.54 / 0.104
  \\
  \multicolumn{1}{c|}{}& \multicolumn{1}{c|}{NAFNet$_A$~\cite{nafnet}}
  & 33.27 / 0.936         
  & \multicolumn{1}{c|}{22.39 / 0.050}          
  & \multicolumn{1}{c|}{NAFNet$_A$~\cite{nafnet}}         
  & 15.97 / 0.597              
  & \multicolumn{1}{c|}{228.6 / 0.454}               
  & \multicolumn{1}{c|}{NAFNet$_A$~\cite{nafnet}}
  & \R{31.42} / \R{0.920}
  & ~2.72 / 0.091
  \\
\multicolumn{1}{c|}{}& \multicolumn{1}{c|}{\textbf{MPerceiver(Ours)}}    
  & 33.40 / 0.937
  & \multicolumn{1}{c|}{\B{17.82} / \B{0.049}}     
  & \multicolumn{1}{c|}{\textbf{MPerceiver(Ours)}}        
  & \B{20.95} / \B{0.644} 
  & \multicolumn{1}{c|}{\B{196.8} / \B{0.437}}   
  &  \multicolumn{1}{c|}{\textbf{MPerceiver(Ours)}}
  & 31.02 / 0.916
  & ~\B{2.31} / \B{0.087}
  \\ 
\multicolumn{1}{c|}{}& \multicolumn{1}{c|}{\textbf{MPerceiver+(Ours)}}    
  & \R{33.69} / \R{0.940}
  & \multicolumn{1}{c|}{\R{17.36} / \R{0.047}}     
  & \multicolumn{1}{c|}{\textbf{MPerceiver+(Ours)}}        
  & \R{21.08} / \R{0.651} 
  & \multicolumn{1}{c|}{\R{190.1} / \R{0.422}}   
  &  \multicolumn{1}{c|}{\textbf{MPerceiver+(Ours)}}
  & \B{31.11} / \B{0.918}
  & ~\R{2.14} / \R{0.085}
  \\ 
  \hline 
  \noalign{\vskip 4pt}
  \hline 
   \multicolumn{1}{c|}{\multirow{+2}*{Type}} & \multicolumn{1}{c|}{\multirow{+2}*{Method}} & \multicolumn{2}{c|}{\textbf{\textit{Raindrop Removal}} (RainDrop)} & 
  \multicolumn{1}{c|}{\multirow{+2}*{Method}} & \multicolumn{2}{c|}{\textbf{\textit{Low-light Enhance.}} (LOL-v2-Real)} & 
  \multicolumn{1}{c|}{\multirow{+2}*{Method}} & \multicolumn{2}{c}{\textbf{\textit{Motion Deblur}} (GoPro)} \\ \cline{3-4} \cline{6-7} \cline{9-10}
  \multicolumn{1}{c|}{\multirow{-2}*{Type}}& \multicolumn{1}{c|}{\multirow{-2}*{Method}} &  ~~~PSNR / SSIM $\uparrow$ &\multicolumn{1}{c|}{~~~~~FID / LPIPS $\downarrow$}& 
  \multicolumn{1}{c|}{\multirow{-2}*{Method}} &  ~~~PSNR / SSIM $\uparrow$ &\multicolumn{1}{c|}{~~~~~FID / LPIPS $\downarrow$}&  
  \multicolumn{1}{c|}{\multirow{-2}*{Method}} &  ~~~PSNR / SSIM $\uparrow$ & ~~~~~FID / LPIPS $\downarrow$               \\ \hline
   \multicolumn{1}{c|}{\multirow{4}{*}{\begin{tabular}[c]{@{}c@{}}Task \\ Specific\end{tabular}}}& \multicolumn{1}{c|}{AttentGAN~\cite{raindrop}} 
  & 31.59 / 0.917          
  & \multicolumn{1}{c|}{{33.33 / 0.056}} 
  & \multicolumn{1}{c|}{SNR~\cite{snr}} 
  & 21.48 / 0.849         
  & \multicolumn{1}{c|}{58.76 / 0.159}              
  & \multicolumn{1}{c|}{MPRNet~\cite{mprnet}}               
  & 32.66 / 0.959
  & 10.98 / 0.091
  \\
  \multicolumn{1}{c|}{}&\multicolumn{1}{c|}{Quan~\etal~\cite{quan2019deep}}
  & 31.37 / 0.918          
  & \multicolumn{1}{c|}{30.56 / 0.065}          
  & \multicolumn{1}{c|}{SNR-SKF~\cite{skf}}         
  & 21.93 / 0.842      
  &  \multicolumn{1}{c|}{73.70 / 0.160}         
  & \multicolumn{1}{c|}{Restormer~\cite{restormer}}
  & 32.92 / 0.961
  & 10.63 / 0.086
  \\
  \multicolumn{1}{c|}{}& \multicolumn{1}{c|}{IDT~\cite{idt}}
  & 31.87 / 0.931         
  & \multicolumn{1}{c|}{25.54 / 0.059}          
  & \multicolumn{1}{c|}{RQ-LLIE~\cite{Liu_2023_ICCV}}            
  & 22.37 / \B{0.854}         
  & \multicolumn{1}{c|}{56.92 / 0.143}               
  & \multicolumn{1}{c|}{Stripformer~\cite{stripformer}}
  & \B{33.08} / \B{0.962}
  & ~\R{9.03} / \R{0.079}
  \\
  \multicolumn{1}{c|}{}& \multicolumn{1}{c|}{UDR-S$^{2}$~\cite{udrs2former}}
  & 32.64 / \B{0.942}          
  & \multicolumn{1}{c|}{27.17 / 0.064}          
  & \multicolumn{1}{c|}{Retinexformer~\cite{Retinexformer}}          
  & \R{22.80} / 0.840        
  & \multicolumn{1}{c|}{62.45 / 0.169}            
  & \multicolumn{1}{c|}{DiffIR~\cite{diffir}}
  & \R{33.20} / \R{0.963}
  & ~\B{9.65} / \B{0.081}
  \\
  \hline
  \multicolumn{1}{c|}{\multirow{7}{*}{\begin{tabular}[c]{@{}c@{}}All \\ in \\One \end{tabular}}} & \multicolumn{1}{c|}{AirNet~\cite{airnet}}
  & 31.32 / 0.925                  
  & \multicolumn{1}{c|}{33.34 / 0.073}          
  & \multicolumn{1}{c|}{AirNet~\cite{airnet}}           
  & 19.69 / 0.821                   
  & \multicolumn{1}{c|}{55.43 / 0.151}              
  & \multicolumn{1}{c|}{AirNet~\cite{airnet}}        
  & 28.31 / 0.910
  & 15.31 / 0.122
  \\

      \multicolumn{1}{c|}{}& \multicolumn{1}{c|}{PromptIR~\cite{promptir}} 
  & 32.03 / 0.938              
  & \multicolumn{1}{c|}{35.75 / 0.073} 
  & \multicolumn{1}{c|}{PromptIR~\cite{promptir}}  
  & 21.23 / \R{0.860}   
  & \multicolumn{1}{c|}{53.92 / 0.145}            
  & \multicolumn{1}{c|}{PromptIR~\cite{promptir}}
  & 31.02 / 0.938
  & 17.54 / 0.131
  \\
      \multicolumn{1}{c|}{}& \multicolumn{1}{c|}{DA-CLIP~\cite{daclip}} 
  & 30.44 / 0.880              
  & \multicolumn{1}{c|}{29.38 / 0.078} 
  & \multicolumn{1}{c|}{DA-CLIP~\cite{daclip}}  
  & 21.76 / 0.762   
  & \multicolumn{1}{c|}{48.23 / 0.134}            
  & \multicolumn{1}{c|}{DA-CLIP~\cite{daclip}}
  & 27.12 / 0.823
  & 16.81 / 0.136
  \\
  \multicolumn{1}{c|}{}& \multicolumn{1}{c|}{Restormer$_A$~\cite{restormer}}
  & 31.75 / 0.936         
  & \multicolumn{1}{c|}{38.22 / 0.075}          
  & \multicolumn{1}{c|}{Restormer$_A$~\cite{restormer}}         
  & 20.77 / 0.851 
  & \multicolumn{1}{c|}{57.04 / 0.155}               
  & \multicolumn{1}{c|}{Restormer$_A$~\cite{restormer}}
  & 30.59 / 0.934 
  & 14.56 / 0.115
  \\
  \multicolumn{1}{c|}{}& \multicolumn{1}{c|}{NAFNet$_A$~\cite{nafnet}}
  & 32.79 / \R{0.943}         
  & \multicolumn{1}{c|}{29.80 / 0.063}          
  & \multicolumn{1}{c|}{NAFNet$_A$~\cite{nafnet}}         
  & 18.04 / 0.827 
  & \multicolumn{1}{c|}{54.25 / 0.147}               
  & \multicolumn{1}{c|}{NAFNet$_A$~\cite{nafnet}}
  & 32.01 / 0.953
  & 13.42 / 0.101
  \\
  \multicolumn{1}{c|}{}& \multicolumn{1}{c|}{\textbf{MPerceiver(Ours)}}    
  & \B{33.21} / 0.929
  & \multicolumn{1}{c|}{\B{21.27} / \B{0.051}}     
  & \multicolumn{1}{c|}{\textbf{MPerceiver(Ours)}}        
  &  22.16 / 0.848     
  & \multicolumn{1}{c|}{\B{45.90} / \B{0.130}}
  &  \multicolumn{1}{c|}{\textbf{MPerceiver(Ours)}}
  & 32.49 / 0.959
  & 10.69 / 0.089
  \\ 
\multicolumn{1}{c|}{}& \multicolumn{1}{c|}{\textbf{MPerceiver+(Ours)}}    
  & \R{33.62} / 0.930
  & \multicolumn{1}{c|}{\R{19.37} / \R{0.044}}     
  & \multicolumn{1}{c|}{\textbf{MPerceiver+(Ours)}}        
  &  \B{22.49} / \B{0.854}     
  & \multicolumn{1}{c|}{\R{45.29} / \R{0.129}}
  &  \multicolumn{1}{c|}{\textbf{MPerceiver+(Ours)}}
  & 32.98 / 0.961
  & 10.51 / 0.087
  \\ 
  \hline
    \noalign{\vskip 4pt}
    \hline
  \multicolumn{1}{c|}{\multirow{+2}*{Type}} & \multicolumn{1}{c|}{\multirow{+2}*{Method}} & \multicolumn{2}{c|}{\textbf{\textit{Defocus Deblur}} (DPDD)} & 
  \multicolumn{1}{c|}{\multirow{+2}*{Method}} & \multicolumn{2}{c|}{\textbf{\textit{Gaussian Denoising}} (Average)} & 
  \multicolumn{1}{c|}{\multirow{+2}*{Method}} & \multicolumn{2}{c}{\textbf{\textit{Real Denoising}} (SIDD)} \\ \cline{3-4} \cline{6-7} \cline{9-10}
  \multicolumn{1}{c|}{\multirow{-2}*{Type}} & \multicolumn{1}{c|}{\multirow{-2}*{Method}} &  ~~~PSNR / SSIM $\uparrow$
  &\multicolumn{1}{c|}{~~~~~FID / LPIPS $\downarrow$}& 
  \multicolumn{1}{c|}{\multirow{-2}*{Method}} &  ~~~PSNR / SSIM $\uparrow$
  &\multicolumn{1}{c|}{~~~~~FID / LPIPS $\downarrow$}& 
  \multicolumn{1}{c|}{\multirow{-2}*{Method}} &  ~~~PSNR / SSIM $\uparrow$ & ~~~~~FID / LPIPS $\downarrow$             \\ \hline
  \multicolumn{1}{c|}{\multirow{4}{*}{\begin{tabular}[c]{@{}c@{}}Task \\ Specific\end{tabular}}}& \multicolumn{1}{c|}{DRBNet~\cite{drbnet}} 
  & 25.73 / 0.791                    
  & \multicolumn{1}{c|}{{49.04 / \B{0.183}}} 
  & \multicolumn{1}{c|}{SwinIR~\cite{swinir}} 
  & 29.60 / 0.842         
  & \multicolumn{1}{c|}{58.99 / 0.146}               
  & \multicolumn{1}{c|}{MPRNet~\cite{mprnet}}               
  & 39.71 / 0.958
  & 49.54 / 0.200
  \\
  \multicolumn{1}{c|}{}& \multicolumn{1}{c|}{Restormer~\cite{restormer}}
  & 25.98 / \R{0.811}          
  & \multicolumn{1}{c|}{\R{44.55} / \R{0.178}}          
    & \multicolumn{1}{c|}{Restormer~\cite{restormer}}            
  & 29.73 / \B{0.845}        
  & \multicolumn{1}{c|}{\B{57.56} / 0.148} 
  & \multicolumn{1}{c|}{Uformer~\cite{uformer}}               
  & 39.89 / \R{0.960}
  & 47.18 / 0.198
  \\
  \multicolumn{1}{c|}{}& \multicolumn{1}{c|}{NRKNet~\cite{nrknet}}
  & \B{26.11} / \B{0.810}          
  & \multicolumn{1}{c|}{55.23 / 0.210}          
  & \multicolumn{1}{c|}{ART~\cite{zhang2022accurate}}         
  & \B{29.79} / \B{0.845}       
  & \multicolumn{1}{c|}{58.50 / \R{0.141}}  
  & \multicolumn{1}{c|}{Restormer~\cite{restormer}}        
  & \B{40.02} / \R{0.960}
  & 47.28 / 0.195
  \\
  \multicolumn{1}{c|}{}& \multicolumn{1}{c|}{FocalNet~\cite{focalnet}}
  & \R{26.18} / 0.808          
  & \multicolumn{1}{c|}{48.82 / 0.210}          
    & \multicolumn{1}{c|}{Xformer~\cite{zhang2023xformer}}          
  & \R{29.83} / \R{0.847}       
  & \multicolumn{1}{c|}{\R{55.22} / \B{0.144}}    
  & \multicolumn{1}{c|}{ART~\cite{zhang2022accurate}}
  & 39.99 / \R{0.960}
  & 42.38 / 0.189
  \\
  \hline
 \multicolumn{1}{c|}{\multirow{7}{*}{\begin{tabular}[c]{@{}c@{}}All \\ in \\One \end{tabular}}} & \multicolumn{1}{c|}{AirNet~\cite{airnet}}
  & 25.37 / 0.770        
  & \multicolumn{1}{c|}{58.82 / 0.193}          
  & \multicolumn{1}{c|}{AirNet~\cite{airnet}}           
  & 28.37 / 0.801          
  & \multicolumn{1}{c|}{69.36 / 0.181}             
  & \multicolumn{1}{c|}{AirNet~\cite{airnet}}
  & 38.32 / 0.945
  & 51.20 / \R{0.134}
  \\
      \multicolumn{1}{c|}{}& \multicolumn{1}{c|}{PromptIR~\cite{promptir}} 
  & 25.66 / 0.791              
  & \multicolumn{1}{c|}{52.64 / 0.197} 
  & \multicolumn{1}{c|}{PromptIR~\cite{promptir}}  
  & 28.82 / 0.816   
  & \multicolumn{1}{c|}{63.76 / 0.170}            
  & \multicolumn{1}{c|}{PromptIR~\cite{promptir}}
  & 39.52 / 0.954
  & 50.52 / 0.198
  \\
      \multicolumn{1}{c|}{}& \multicolumn{1}{c|}{DA-CLIP~\cite{daclip}} 
  & 24.91 / 0.749              
  & \multicolumn{1}{c|}{57.43 / 0.201} 
  & \multicolumn{1}{c|}{DA-CLIP~\cite{daclip}}  
  & 25.13 / 0.692   
  & \multicolumn{1}{c|}{59.82 / 0.235}            
  & \multicolumn{1}{c|}{DA-CLIP~\cite{daclip}}
  & 34.04 / 0.824
  & \R{34.56} / \B{0.186}
  \\
   \multicolumn{1}{c|}{}&\multicolumn{1}{c|}{Restormer$_A$~\cite{restormer}}
  & 25.74 / 0.795         
  & \multicolumn{1}{c|}{54.74 / 0.213}          
  & \multicolumn{1}{c|}{Restormer$_A$~\cite{restormer}}         
  & 28.65 / 0.812
  & \multicolumn{1}{c|}{63.48 / 0.172}              
  & \multicolumn{1}{c|}{Restormer$_A$~\cite{restormer}}
  & 39.48 / 0.954
  & 51.75 / 0.190
  \\
  \multicolumn{1}{c|}{}&\multicolumn{1}{c|}{NAFNet$_A$~\cite{nafnet}}
  & 25.85 / 0.803         
  & \multicolumn{1}{c|}{48.45 / 0.191}          
  & \multicolumn{1}{c|}{NAFNet$_A$~\cite{nafnet}}         
  & 29.21 / 0.829 
  & \multicolumn{1}{c|}{60.84 / 0.163}              
  & \multicolumn{1}{c|}{NAFNet$_A$~\cite{nafnet}}
  & 39.76 / 0.957
  & 45.54 / 0.197
  \\
\multicolumn{1}{c|}{}& \multicolumn{1}{c|}{\textbf{MPerceiver(Ours)}}    
  & 25.88 / 0.803
  & \multicolumn{1}{c|}{48.22 / 0.190}     
  & \multicolumn{1}{c|}{\textbf{MPerceiver(Ours)}}       
  & 29.57 / 0.838     
  & \multicolumn{1}{c|}{61.44 / 0.158}   
  &  \multicolumn{1}{c|}{\textbf{MPerceiver(Ours)}}
  & 39.96 / \B{0.959}
  & \B{41.11} / 0.191
  \\ 
\multicolumn{1}{c|}{}& \multicolumn{1}{c|}{\textbf{MPerceiver+(Ours)}}    
  & 26.06 / 0.805
  & \multicolumn{1}{c|}{\B{46.07} / 0.190}     
  & \multicolumn{1}{c|}{\textbf{MPerceiver+(Ours)}}       
  & 29.61 / 0.839      
  & \multicolumn{1}{c|}{60.91 / 0.156}   
  &  \multicolumn{1}{c|}{\textbf{MPerceiver+(Ours)}}
  & \R{40.05} / \R{0.960}
  & 41.46 / 0.190
  \\ 
  \hline  
  \bottomrule
  \end{tabular}
  \vspace{-4mm}
\end{table*}
\begin{table*}[thb] \footnotesize
\centering
    \caption{\textbf{[All-in-one]} Quantitative comparison on the proposed mixed degradation benchmark MID6. }
    \vspace{-3mm}
    \label{tab:compare_mid}
    \setlength{\tabcolsep}{1.9pt}
    \resizebox{\textwidth}{!}{
        \begin{tabular}{c|*{3}{c}|*{3}{c}|*{3}{c}|*{3}{c}|*{3}{c}|*{3}{c}}
        \toprule
        \multirow{2}{*}{Method} & \multicolumn{3}{c|}{Haze \&Noise \&Blur} 
                               & \multicolumn{3}{c|}{Lowlight \&Noise \&Blur} 
                               & \multicolumn{3}{c|}{Rain \&Noise \&Blur} 
                               & \multicolumn{3}{c|}{Rain \&Raindrop \&Noise}  
                               & \multicolumn{3}{c|}{Raindrop \&Noise \&Blur}
                               & \multicolumn{3}{c}{Snow \&Noise \&Blur} \\ \cline{2-19}
                               & PSNR $\uparrow$ & SSIM $\uparrow$ & LPIPS $\downarrow$
                               & PSNR $\uparrow$ & SSIM $\uparrow$ & LPIPS $\downarrow$
                               & PSNR $\uparrow$ & SSIM $\uparrow$ & LPIPS $\downarrow$  
                               & PSNR $\uparrow$ & SSIM $\uparrow$ & LPIPS $\downarrow$
                               & PSNR $\uparrow$ & SSIM $\uparrow$ & LPIPS $\downarrow$
                               & PSNR $\uparrow$ & SSIM $\uparrow$ & LPIPS $\downarrow$
                               \\
        \midrule
        AirNet~\cite{airnet} & 17.51 & 0.613 & 0.440 & 17.09 & 0.552 & 0.527 & 24.31 & 0.601 & 0.364 & 21.46 & \B{0.570} & 0.471 & 26.44 & 0.745 & 0.353 & 22.62 &0.566 & 0.454  \\
        TransWeather~\cite{valanarasu2022transweather} & \B{25.11} & \B{0.739} & \B{0.241} & 20.36 & 0.626 & \B{0.408} & \B{25.01} & \B{0.683} & \B{0.294}   & 21.59 & 0.541 & \B{0.412} & \B{27.76} & 0.737 & \B{0.229} & \B{23.90} & \B{0.672} & \B{0.306}\\
        WGWS-Net~\cite{zhu2023learning} & 17.66 & 0.617 & 0.394 & 17.57 & 0.570 & 0.448 & 22.10 & 0.600 & 0.353   & 20.12 & 0.522 & 0.446 & 25.67 & 0.718 & 0.307 & 20.03 & 0.569 & 0.401  \\
        PromptIR~\cite{promptir} & 18.41 & 0.631 & 0.437 & \B{20.95} & \B{0.649} & 0.413 & 23.75 & 0.647 & 0.313   & 21.31 & 0.556 & 0.461 & 25.41 & 0.721 & 0.327 & 20.92 & 0.601 & 0.362  \\
        Restormer$_A$~\cite{restormer} & 17.03 & 0.602 & 0.470 & 16.49 & 0.541 & 0.498 & 23.22 & 0.611 & 0.332 & 20.39 & 0.561 & 0.493 & 24.48 & 0.697 & 0.401 & 21.39 & 0.606 & 0.392  \\
        NAFNet$_A$~\cite{nafnet} & 16.59 & 0.548 & 0.541 & 15.72 & 0.605  & 0.520 & 23.48 & 0.563 & 0.346 & \R{22.72} & \R{0.599} & 0.439 & 27.20 & \B{0.769} & 0.307 & 20.28 & 0.535 & 0.484  \\
        \textbf{MPerceiver} (Ours) & \R{26.19}  & \R{0.782} & \R{0.211} & \R{23.84} & \R{0.671} & \R{0.343} & \R{26.00} & \R{0.762} & \R{0.193}   & \B{22.35} & 0.525 & \R{0.268} & \R{28.49} & \R{0.771} & \R{0.127} & \R{24.36} & \R{0.719} & \R{0.263} \\ 
        \bottomrule
    \end{tabular}
        }
    \vspace{-3.5mm}
\end{table*}

\begin{table*}[t]
  \caption{\textbf{[All-in-one]} Quantitative comparison on real-world datasets of \textbf{\textit{deraining}}, \textbf{\textit{desnowing}} and \textbf{\textit{motion deblurring}}. SSID~\cite{ssid} has no GT images. Methods are directly applied to the LHP~\cite{lhp} and RealBlur-J~\cite{realblur} sets to evaluate generalization to real-world images. MUSS~\cite{ssid} is a semi-supervised deraining model trained with additional real rainy data, denoted with * for reference.}
  \vspace{-3mm}
 \label{tab:compare_real}
  \centering
  \renewcommand\arraystretch{0.88}
  \fontsize{6.6pt}{\baselineskip}\selectfont
  \setlength\tabcolsep{1.4pt}
  \begin{tabular}{ccccccccccccc}
  \toprule
  \hline
 \multicolumn{1}{c|}{\multirow{+2}*{Type}}& \multicolumn{3}{c|}{\textbf{\textit{Deraining}} (LHP~\cite{lhp})}  & \multicolumn{3}{c|}{\textbf{\textit{Deraining}} (SSID~\cite{ssid})} & 
 \multicolumn{3}{c|}{\textbf{\textit{Desnowing}} (RealSnow~\cite{zhu2023learning})} &\multicolumn{3}{c}{\textbf{\textit{Motion Deblur}} (RealBlur-J~\cite{realblur}))} \\ \cline{2-13}
  \multicolumn{1}{c|}{\multirow{-2}*{Type}}&  \multicolumn{1}{c|}{Method} &  ~~~PSNR $\uparrow$ & \multicolumn{1}{c|}{~~~SSIM $\uparrow$}& 
   \multicolumn{1}{c|}{Method} &  ~~~NIQE $\downarrow$ & \multicolumn{1}{c|}{~BRISQUE $\downarrow$}& 
  \multicolumn{1}{c|}{Method} & ~~~PSNR $\uparrow$ & \multicolumn{1}{c|}{~~~SSIM $\uparrow$} & \multicolumn{1}{c|}{Method} & ~~~PSNR $\uparrow$ & ~~~SSIM $\uparrow$     \\ \hline
   \multicolumn{1}{c|}{\multirow{4}{*}{\begin{tabular}[c]{@{}c@{}}Task \\ Specific\end{tabular}}} & \multicolumn{1}{c|}{MUSS*~\cite{ssid}} 
  & 30.02          
  & \multicolumn{1}{c|}{{0.886}} 
  & \multicolumn{1}{c|}{MUSS*~\cite{ssid}} 
  & \R{3.43}        
  & \multicolumn{1}{c|}{\R{28.97}}               
  & \multicolumn{1}{c|}{MIRNetv2~\cite{mirnetv2}}               
  & 31.39
  & \multicolumn{1}{c|}{0.916}
  & \multicolumn{1}{c|}{MPRNet~\cite{mprnet}}
  & 28.70 
  & 0.873
  \\
  \multicolumn{1}{c|}{}& \multicolumn{1}{c|}{Restormer~\cite{restormer}}
  & 29.72          
  & \multicolumn{1}{c|}{\B{0.889}}          
  & \multicolumn{1}{c|}{Restormer~\cite{restormer}}          
  & 4.12       
  & \multicolumn{1}{c|}{33.29}               
  & \multicolumn{1}{c|}{ART~\cite{zhang2022accurate}} 
  & 31.05
  & \multicolumn{1}{c|}{0.913}
  & \multicolumn{1}{c|}{Restormer~\cite{restormer}} 
  & 28.96
  & 0.879
  \\
  \multicolumn{1}{c|}{}& \multicolumn{1}{c|}{DRSformer~\cite{drsformer}}
  & 30.04        
  &  \multicolumn{1}{c|}{\R{0.895}}      
  & \multicolumn{1}{c|}{DRSformer~\cite{drsformer}}            
  & 4.19                        
  & \multicolumn{1}{c|}{35.52}               
  & \multicolumn{1}{c|}{Restormer~\cite{restormer}}     
  & 31.38
  & \multicolumn{1}{c|}{\B{0.923}}
  & \multicolumn{1}{c|}{Stripformer~\cite{stripformer}}     
  & 28.82
  & 0.876
  \\
  \multicolumn{1}{c|}{}& \multicolumn{1}{c|}{UDR-S$^{2}$~\cite{udrs2former}}
  & 28.59          
  & \multicolumn{1}{c|}{0.884}          
  & \multicolumn{1}{c|}{UDR-S$^{2}$~\cite{udrs2former}}         
  & 3.77               
  & \multicolumn{1}{c|}{35.86}              
  & \multicolumn{1}{c|}{NAFNet~\cite{nafnet}}
  & \B{31.44}
  & \multicolumn{1}{c|}{0.919}
  & \multicolumn{1}{c|}{DiffIR~\cite{diffir}}
  & \B{29.06}
  & \R{0.882}
  \\
  \hline
  \multicolumn{1}{c|}{\multirow{4}{*}{\begin{tabular}[c]{@{}c@{}}All \\ in \\One \end{tabular}}} & \multicolumn{1}{c|}{AirNet~\cite{airnet}}
  & \B{31.73}                     
  & \multicolumn{1}{c|}{\B{0.889}}          
  & \multicolumn{1}{c|}{AirNet~\cite{airnet}}           
  & 3.69                    
  & \multicolumn{1}{c|}{ 30.91}              
  & \multicolumn{1}{c|}{AirNet~\cite{airnet}}
  & 31.02
  & \multicolumn{1}{c|}{\B{0.923}}
    & \multicolumn{1}{c|}{AirNet~\cite{airnet}}
  & 27.91
  & 0.834
  \\
  \multicolumn{1}{c|}{}& \multicolumn{1}{c|}{TransWeather~\cite{valanarasu2022transweather}} 
  & 29.87                   
  & \multicolumn{1}{c|}{0.867}          
  & \multicolumn{1}{c|}{TransWeather~\cite{valanarasu2022transweather}}          
  & 3.96
  & \multicolumn{1}{c|}{30.94}            
  & \multicolumn{1}{c|}{TransWeather~\cite{valanarasu2022transweather}}
  & 31.13
  & \multicolumn{1}{c|}{0.922}
    & \multicolumn{1}{c|}{TransWeather~\cite{valanarasu2022transweather}}
  & 28.03
  & 0.837
  \\
  \multicolumn{1}{c|}{}& \multicolumn{1}{c|}{WGWS-Net~\cite{zhu2023learning}} 
  & 30.77              
  & \multicolumn{1}{c|}{0.885} 
  & \multicolumn{1}{c|}{WGWS-Net~\cite{zhu2023learning}}  
  & 3.71
  & \multicolumn{1}{c|}{30.79}            
  & \multicolumn{1}{c|}{WGWS-Net~\cite{zhu2023learning}}
  & 31.37 
  & \multicolumn{1}{c|}{0.919}
    & \multicolumn{1}{c|}{WGWS-Net~\cite{zhu2023learning}}
  & 28.10
  & 0.838
  \\
\multicolumn{1}{c|}{}& \multicolumn{1}{c|}{\textbf{MPerceiver} (Ours)}    
  & \R{32.07}
  & \multicolumn{1}{c|}{\B{0.889}}     
  & \multicolumn{1}{c|}{\textbf{MPerceiver} (Ours)}        
  & \B{3.60} 
  & \multicolumn{1}{c|}{\B{30.77}}   
  &  \multicolumn{1}{c|}{\textbf{MPerceiver} (Ours)}
  & \R{31.45}
  & \multicolumn{1}{c|}{\R{0.924}}
    &  \multicolumn{1}{c|}{\textbf{MPerceiver} (Ours)}
  & \R{29.13}
  & \B{0.881}
  \\ 
  \hline 
    \bottomrule
  \end{tabular}
  \vspace{-6mm}
\end{table*}
\begin{table}[thb]\centering 
    \caption{\textbf{[Zero-shot]} \textbf{\textit{UDC IR}} (TOLED / POLED) results.} %
        \vspace{-3mm}
    \scriptsize
    \label{tab:zs_udc}
    \setlength{\tabcolsep}{3.0pt}
    \resizebox{0.48\textwidth}{!}{
    \begin{tabular}{*{1}{c}|*{3}{c}|*{3}{c}}
        \toprule
         \multicolumn{1}{c|}{\multirow{2}{*}{Method}} & \multicolumn{3}{c|}{TOLED~\cite{udc}} & \multicolumn{3}{c}{POLED~\cite{udc}}\\ \cline{2-7}
          & PSNR $\uparrow$ & SSIM $\uparrow$ & LPIPS $\downarrow$ & PSNR $\uparrow$ & SSIM $\uparrow$  & LPIPS $\downarrow$ \\
        \midrule
         AirNet~\cite{airnet}  & 26.76 & 0.799 & 0.307 &  13.49 & 0.522 & 0.696 \\ %
        TransWeather~\cite{valanarasu2022transweather}  & 27.58 &  0.810  & 0.316 & \B{15.86} & \B{0.590} & 0.707\\
        WGWS-Net~\cite{zhu2023learning}  & 22.11  & 0.731 & 0.374 & 10.96 & 0.429 & 0.776\\
        Restormer$_A$~\cite{restormer} & 27.74 & 0.841   & \B{0.294} & 13.94 & 0.528 & \B{0.681} \\
        NAFNet$_A$~\cite{nafnet} & \B{27.90} & \B{0.848} & 0.320 & 10.68 & 0.555 & 0.713 \\
        \textbf{MPerceiver} (Ours) & \R{32.92} & \R{0.863} &\R{0.161} & \R{20.41} & \R{0.650} & \R{0.445} \\
        \bottomrule
    \end{tabular}
    }
    \vspace{-4mm}
\end{table}

\begin{table}[thb]\centering 
    \caption{\textbf{[Zero-shot]} \textbf{\textit{Underwater IR}} results.} %
        \vspace{-3mm}
    \scriptsize
    \label{tab:zs_uir}
    \setlength{\tabcolsep}{3.0pt}
    \resizebox{0.48\textwidth}{!}{
    \begin{tabular}{*{1}{c}|*{3}{c}|*{3}{c}}
        \toprule
         \multicolumn{1}{c|}{\multirow{2}{*}{Method}} & \multicolumn{3}{c|}{UIEB~\cite{uieb}} & \multicolumn{3}{c}{UWCNN\cite{Anwar2019UWCNN}}\\ \cline{2-7}
          & PSNR $\uparrow$ & SSIM $\uparrow$ & LPIPS $\downarrow$ & PSNR $\uparrow$ & SSIM $\uparrow$  & LPIPS $\downarrow$ \\
        \midrule
         AirNet~\cite{airnet}  & 17.09 & 0.761 & 0.304 &  13.54 & 0.737 & 0.403 \\ %
        TransWeather~\cite{valanarasu2022transweather}  & 17.17 &  0.754  & 0.303 & 13.59 & 0.731  & 0.408\\
        WGWS-Net~\cite{zhu2023learning}  & 16.99  & 0.745 & 0.340 & \B{13.83} & \B{0.740} & 0.410\\
        Restormer$_A$~\cite{restormer} & \B{17.34} & \B{0.770} & \B{0.300} & 13.49 & 0.737 & \B{0.401} \\
        NAFNet$_A$~\cite{nafnet} & 17.31 & 0.736 & 0.307 & 13.62 & 0.736 & 0.405 \\
        \textbf{MPerceiver} (Ours) & \R{22.69} & \R{0.902} &\R{0.150} & \R{14.77} & \R{0.774} & \R{0.299} \\
        \bottomrule
    \end{tabular}
    }
    \vspace{-4mm}
\end{table}

\begin{table}[thb]\centering
    \caption{\textbf{[Few-shot]} \textbf{\textit{Color JPEG compression artifact removal}} (QF=10) results. We only use 100 images from DIV2K to fine-tune all-in-one methods, while task-specific methods adopt DIV2K and Flickr2K as the training set (3450 images). } %
    \label{tab:compare_fs_jpeg}
        \scriptsize
    \vspace{-3mm}
    \resizebox{0.48\textwidth}{!}{
    \begin{tabular}{*{1}{c}|*{1}{c}|*{2}{c}|*{2}{c}}
        \toprule
         \multicolumn{1}{c|}{\multirow{2}{*}{Type}} & \multicolumn{1}{c|}{\multirow{2}{*}{Method}} & \multicolumn{2}{c|}{LIVE1~\cite{sheikh2005live}} & \multicolumn{2}{c}{BSD500~\cite{bsd}}\\ \cline{3-6}
         & & PSNR $\uparrow$ & SSIM $\uparrow$ & PSNR $\uparrow$ & SSIM $\uparrow$ \\
         \midrule
         \multirow{2}{*}{\begin{tabular}[c]{@{}c@{}}Task \\ Specific\end{tabular}} & QGAC~\cite{qgac} & 27.62 & \R{0.804} & 27.74 & \R{0.802} \\
         & FBCNN~\cite{fbcnn} & \B{27.77} & \B{0.803} & \B{27.85} & \B{0.799} \\
        \midrule
         \multirow{4}{*}{All-in-One} & AirNet~\cite{airnet}  & 27.47 & 0.797 &  27.60 & 0.788 \\ %
        & TransWeather~\cite{valanarasu2022transweather}  & 26.45 &  0.755 & 26.68 & 0.785 \\
        & WGWS-Net~\cite{zhu2023learning}  & 26.50  & 0.750 & 26.60 & 0.741 \\
        & \textbf{MPerceiver} (Ours) & \R{27.79} & \R{0.804} & \R{27.88} & 0.795  \\
        \bottomrule
    \end{tabular}
    }
    \vspace{-4mm}
\end{table}

\begin{table}[!t]
\begin{center}
\caption{\textbf{[Few-shot]} \textbf{\textit{Image demosaicking}} results. The training setting is the same as JPEG compression artifact removal.}
\label{table:fs_mosaick}
\vspace{-3mm}
\setlength{\tabcolsep}{1.9pt}
\resizebox{0.48\textwidth}{!}{
\begin{tabular}{l|ccc|cccc}
\toprule
Datasets	&\makecell{RLDD \\ \cite{guo2020residual}}	&\makecell{RNAN \\ \cite{zhang2019rnan}}	&\makecell{DRUNet \\ \cite{zhang2021plug}} &\makecell{AirNet \\ \cite{airnet}} &\makecell{TransWeather \\ \cite{valanarasu2022transweather}} &\makecell{WGWS-Net \\ \cite{zhu2023learning}}  &\makecell{\textbf{MPerceiver} \\ (Ours)}	\\ \midrule
Kodak~\cite{franzen1999kodak}	&42.49	&\R{43.16}	&42.68 & 40.55 & 39.58 & 41.22 & \B{43.06}		\\
McMaster~\cite{zhang2011McMaster}&39.25	&\R{39.70}	&39.39 & 37.36 & 36.68 & 38.06 & \B{39.68}	\\
\bottomrule[0.1em]
\end{tabular}}
\end{center}
\vspace{-7mm}
\end{table}
\begin{table}[thb]\centering
    \caption{\textbf{[Few-shot]} Quantitative comparison on the \textbf{\textit{demoireing}} dataset TIP2018~\cite{tip18}. Note that we only use 5\% of the training data to fine-tune pre-trained all-in-one models. } %
    \label{tab:compare_fs_moire}
        \scriptsize
    \vspace{-2mm}
    \resizebox{0.48\textwidth}{!}{
    \begin{tabular}{*{1}{c}|*{1}{c}|*{1}{c}|*{2}{c}}
        \toprule
         \multicolumn{1}{c|}{Type} & \multicolumn{1}{c|}{Method} & \multicolumn{1}{c|}{Venue} & PSNR $\uparrow$ & SSIM $\uparrow$ \\
         \midrule
         \multirow{4}{*}{\begin{tabular}[c]{@{}c@{}}Task \\ Specific\end{tabular}} & MBCNN~\cite{mbcnn} & \textit{CVPR $^\prime$ 20} & 30.03 &  0.893\\
         & FHDe$^2$Net~\cite{he2020fhde} & \textit{ECCV $^\prime$ 20} & 27.78 &  0.896\\
         & WDNet~\cite{wdnet} & \textit{ECCV $^\prime$ 20} & 28.08 &  \B{0.904}\\
         & ESDNet~\cite{esdnet} & \textit{ECCV $^\prime$ 22} & \B{30.11} &  \R{0.920}\\
         \midrule
         \multirow{4}{*}{All-in-One} & AirNet~\cite{airnet} & \textit{CVPR $^\prime$ 22} & 28.59 &  0.866\\
         & TransWeather~\cite{valanarasu2022transweather} & \textit{CVPR $^\prime$ 22} & 27.68 &  0.848 \\
         & WGWS-Net~\cite{zhu2023learning} & \textit{CVPR $^\prime$ 23} & 28.13 &  0.861 \\
         & \textbf{MPerceiver} (Ours) & - & \R{30.19} &  0.885\\
        \bottomrule
    \end{tabular}
    }
    \vspace{-4mm}
\end{table}

\subsection{Experimental Setup}
\textbf{Settings.}
\textbf{(1) All-in-one:} We train a unified model to solve 10 IR tasks, including deraining, dehazing, desnowing, raindrop removal, low-light enhancement, motion deblurring, defocus deblurring, gaussian denoising, real denoising and challenging mixed degradations removal.  \textbf{(2) Zero-shot:} We use the all-in-one pre-trained model to directly solve training-unseen tasks, including under-display camera IR (POLED/TOLED), underwater IR. \textbf{(3) Few-shot:} We fine-tune the all-in-one pre-trained model using a small amount of data (about 3\%-5\% of the data used by task-specific methods) and adapt it to new tasks, including JPEG compression artifact removal, demosaicking, demoireing.

\noindent \textbf{Datasets and Metrics.}
For \textbf{setting(1)}, a combination of various image degradation datasets is used to evaluate our method, \ie, Rain1400~\cite{rain1400}, Outdoor-Rain~\cite{outdoor}, SSID~\cite{ssid} and LHP~\cite{lhp} for deraining; RESIDE~\cite{reside}, NH-HAZE~\cite{nh} and Dense-Haze~\cite{dense} for dehazing; Snow100K~\cite{snow100k} and RealSnow~\cite{zhu2023learning} for desnowing; RainDrop~\cite{raindrop} and RainDS~\cite{rainds} for raindrop removal; LOL-v2~\cite{lolv2} for low-light enhancement; CBSD68, CBSD400~\cite{bsd}, Urban100~\cite{urban100}, Kodak24~\cite{franzen1999kodak}, McMaster~\cite{zhang2011McMaster}, WED~\cite{wed} and DF2K for gaussian denoising; SIDD~\cite{sidd} for real image denoising; GoPro~\cite{gopro} and RealBlur~\cite{realblur} for motion deblurring; and DPDD~\cite{dpdd} for defocus deblurring. Considering real-world LQ images may contain more than just a single degradation, we construct a challenging mixed degradation benchmark named MID6, in which the LQ images contain mixed degradations (\eg, low-light\&noise\&blur, rain\&raindrop\&noise; See Fig.~\ref{fig:visual_result_mid}). For \textbf{setting(2)}, we utilize TOLED~\cite{udc} and POLED~\cite{udc} for under-display camera (UDC) IR. Following~\cite{huang2023contrastive}, 90 pairs from UIEB~\cite{uieb} and 110 pairs from UWCNN~\cite{Anwar2019UWCNN} are used for underwater IR. For \textbf{setting(3)}, we use the first 100 images of DIV2K~\cite{DIV2K} to fine-tune our method for JPEG compression artifact removal and demosaicking. 5\% of the data in TIP2018~\cite{tip18} training set is used to fine-tune our method for demoireing. We provide a detailed introduction to the datasets used for training and testing in the Appendix.

We adopt PSNR and SSIM as the distortion metrics, LPIPS~\cite{lpips} and FID~\cite{fid} as the perceptual metrics, NIQE~\cite{niqe} and BRISQUE~\cite{brisque} as no-reference metrics.

\noindent\textbf{Implementation Details.} 
We use the AdamW optimizer ($\beta_1=0.9$, $\beta_2=0.999$) with the initial learning rate $1e^{-4}$ gradually reduced to $1e^{-6}$ with the cosine annealing~\cite{loshchilov2016sgdr} schedule to train our model. The training runs for 500 epochs on 8 NVIDIA A100 GPUs. We adopt DDIM~\cite{ddim} as our sampling strategy (50 steps). Our model is based on SD 2.1. More details are presented in the Appendix.

\subsection{Comparison with state-of-the-art methods}
\textbf{All-in-one.} We conduct comparisons between our method and SOTA all-in-one methods as well as task-specific methods. To ensure a fair evaluation, we train all-in-one models from scratch employing our training strategy. Given that Restormer~\cite{restormer} and NAFNet~\cite{nafnet} serve as strong general IR baselines, we additionally train both a Restormer and a NAFNet model within the all-in-one setting.

\begin{figure*}[!htbp]
	\captionsetup{font=small}
	\scriptsize
	\centering
	
	\newcommand{\h}{0.105}
	\newcommand{\wa}{0.12}
	\newcommand{\wb}{0.16}
	\newcommand{\g}{-0.7mm}

 	\setlength\tabcolsep{1.8pt}
	\renewcommand{\arraystretch}{1}
	\resizebox{1.00\linewidth}{!} {
			\renewcommand{\h}{0.15}
			\newcommand{\w}{0.200}
				\begin{adjustbox}{valign=t}
					\begin{tabular}{cccccccc}
						\includegraphics[height=\h \textwidth, width=\w \textwidth]{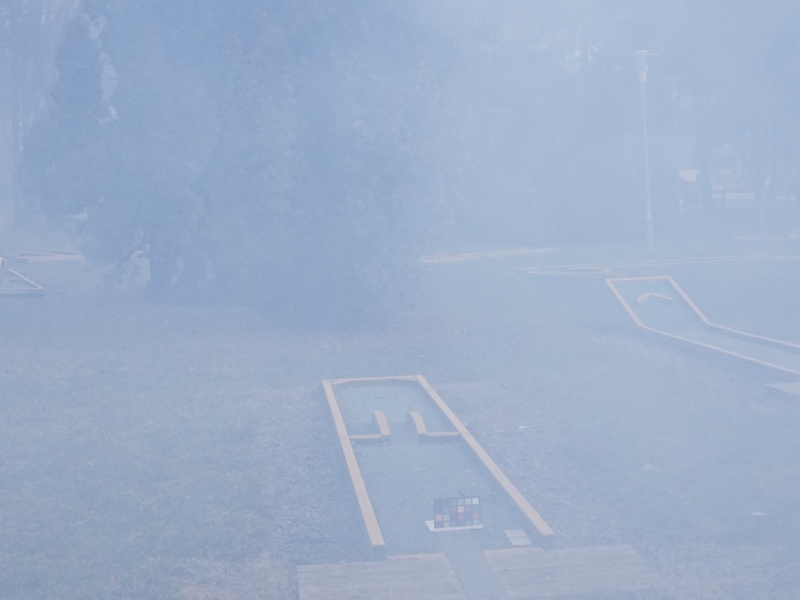} \hspace{\g} &
						\includegraphics[height=\h \textwidth, width=\w \textwidth]{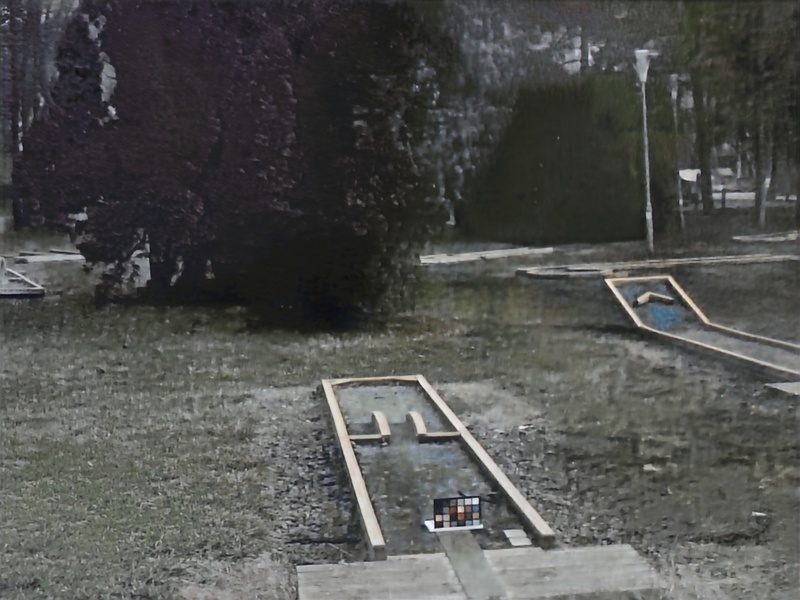} \hspace{\g} &
						\includegraphics[height=\h \textwidth, width=\w \textwidth]{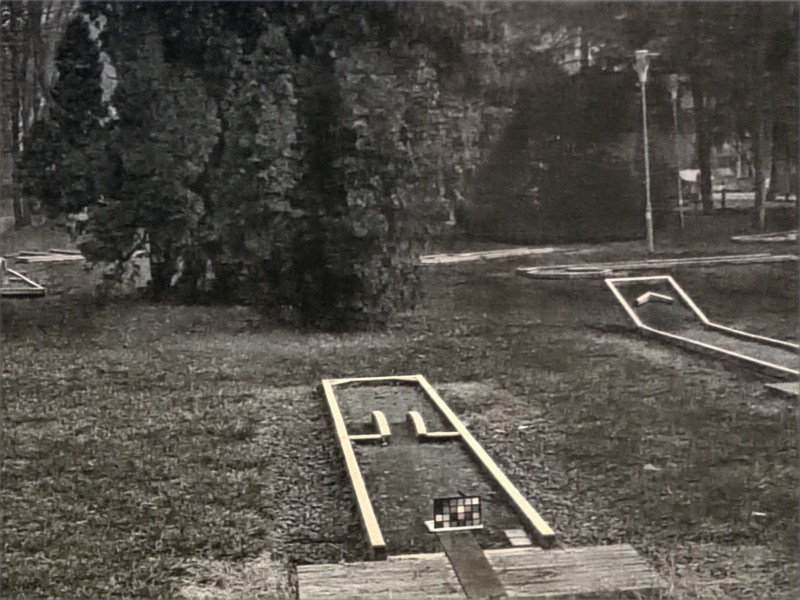} \hspace{\g} &
                        \includegraphics[height=\h \textwidth, width=\w \textwidth]{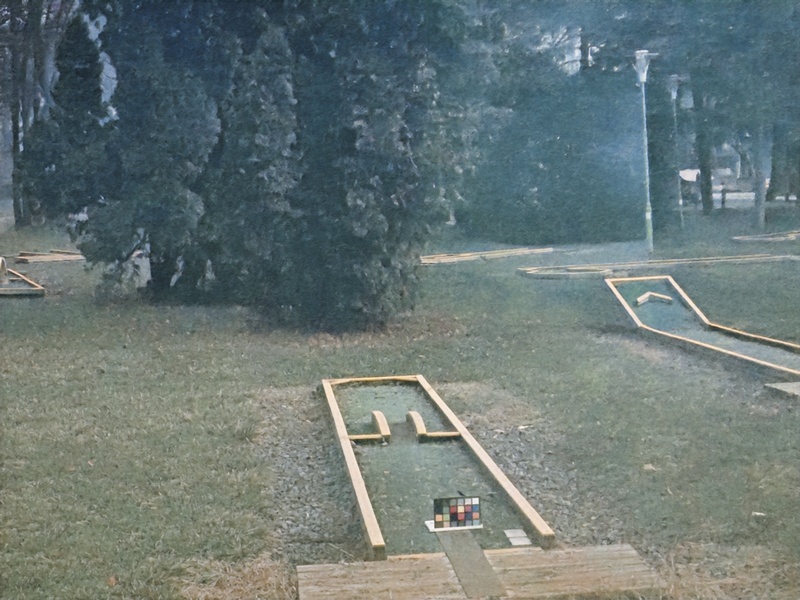} \hspace{\g} &
                        \includegraphics[height=\h \textwidth, width=\w \textwidth]{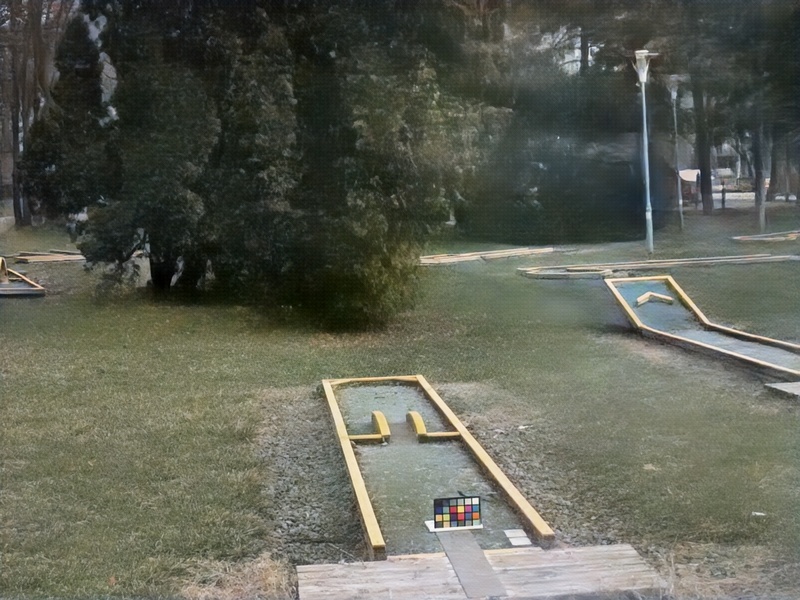} \hspace{\g} &
                        \includegraphics[height=\h \textwidth, width=\w \textwidth]{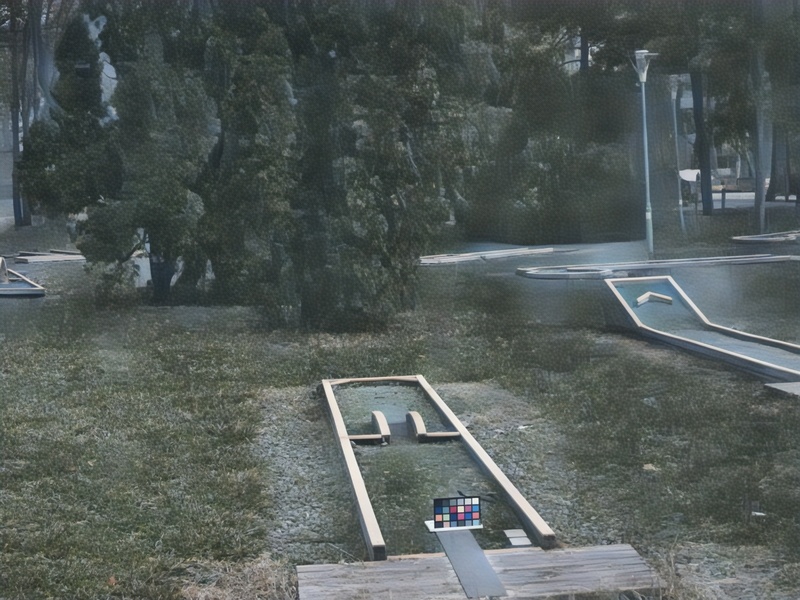} \hspace{\g} &
                        \includegraphics[height=\h \textwidth, width=\w \textwidth]{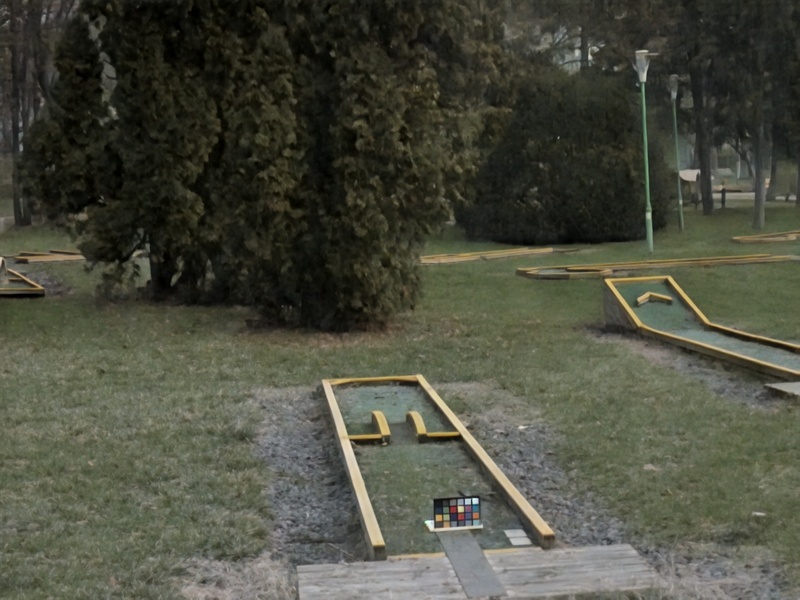} \hspace{\g} &
                        \includegraphics[height=\h \textwidth, width=\w \textwidth]{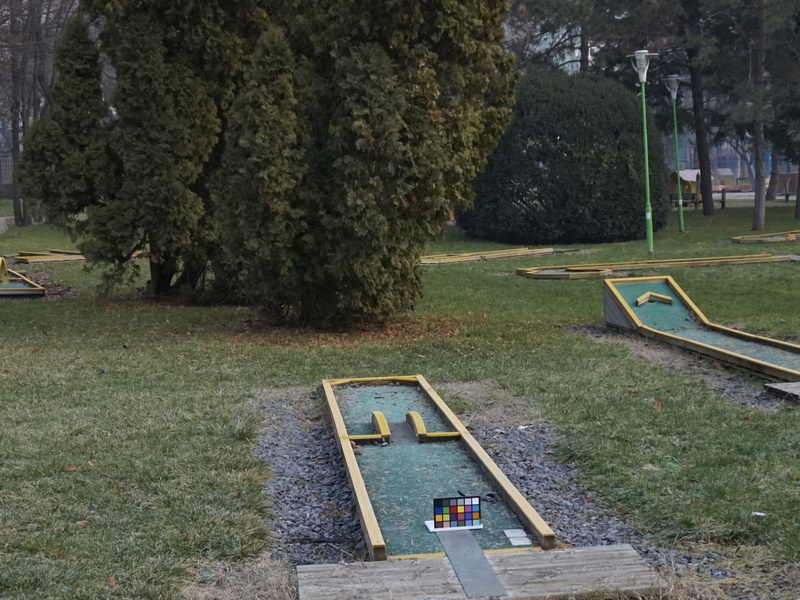} 
						\\
						Hazy Input / 7.10 dB \hspace{\g} &
						Dehamer~\cite{dehamer} / 22.83 dB \hspace{\g} & 
                        MB-Taylor~\cite{taylorformer} / 21.65 dB \hspace{\g} & 
                        AirNet~\cite{airnet} / 15.03 dB \hspace{\g} & 
                        TransWeather~\cite{valanarasu2022transweather} / 19.94 dB \hspace{\g} & 
                        WGWS-Net~\cite{zhu2023learning} / 19.98 dB \hspace{\g} & 
                        \textbf{Ours} / 25.03 dB \hspace{\g} & 
                        GT / PSNR 
						\\
					\end{tabular}
				\end{adjustbox}
		 }

    	\setlength\tabcolsep{1.8pt}
	\renewcommand{\arraystretch}{1}
	\resizebox{1.00\linewidth}{!} {
			\renewcommand{\h}{0.15}
			\newcommand{\w}{0.200}
				\begin{adjustbox}{valign=t}
					\begin{tabular}{cccccccc}
						\includegraphics[height=\h \textwidth, width=\w \textwidth]{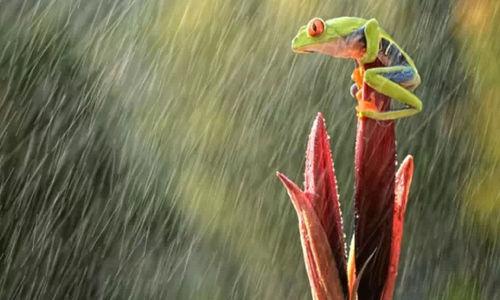} \hspace{\g} &
						\includegraphics[height=\h \textwidth, width=\w \textwidth]{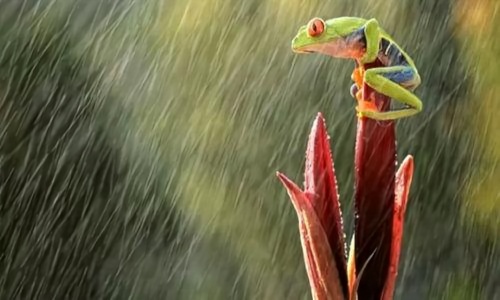} \hspace{\g} &
                        \includegraphics[height=\h \textwidth, width=\w \textwidth]{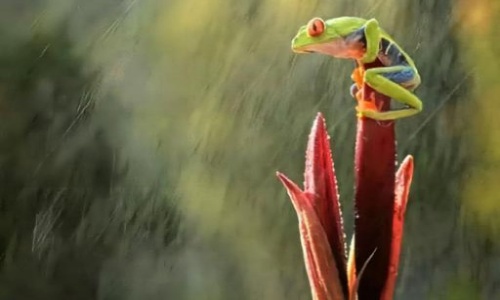} \hspace{\g} &
                        \includegraphics[height=\h \textwidth, width=\w \textwidth]{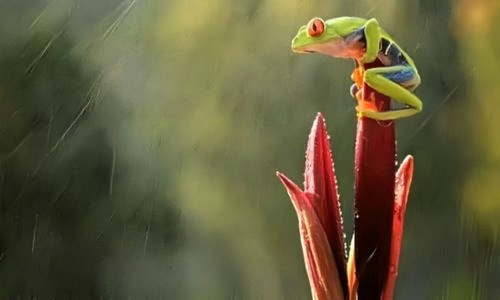} \hspace{\g} &
                        \includegraphics[height=\h \textwidth, width=\w \textwidth]{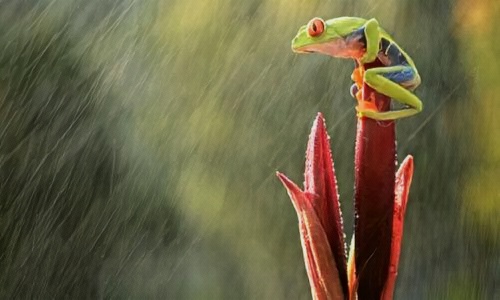} \hspace{\g} &
                        \includegraphics[height=\h \textwidth, width=\w \textwidth]{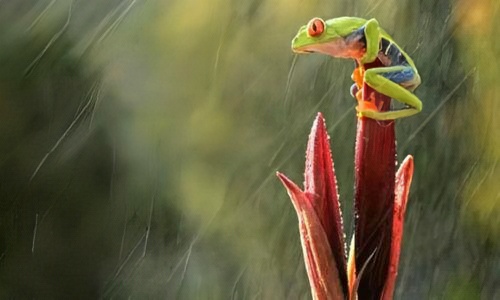} \hspace{\g} &
                        \includegraphics[height=\h \textwidth, width=\w \textwidth]{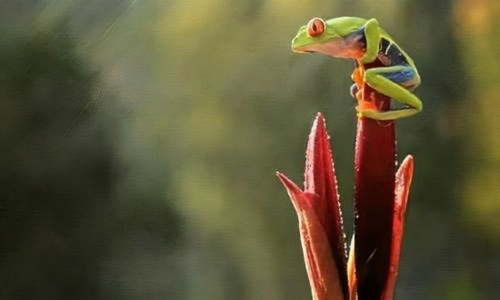} \hspace{\g} &
                        \includegraphics[height=\h \textwidth, width=\w \textwidth]{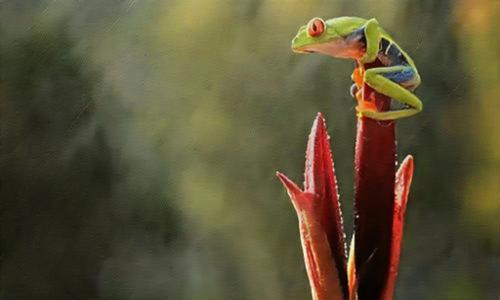}
						\\
						Rainy Input \hspace{\g} &
                        DRSformer~\cite{drsformer}  \hspace{\g} & 
                        UDR-S$^2$~\cite{udrs2former}  \hspace{\g} & 
                        AirNet~\cite{airnet}  \hspace{\g} & 
                        TransWeather~\cite{valanarasu2022transweather}  \hspace{\g} & 
                        WGWS-Net~\cite{zhu2023learning} \hspace{\g} & 
                        \textbf{Ours} \hspace{\g} & 
                        MUSS*~\cite{ssid}
						\\
					\end{tabular}
				\end{adjustbox}
		 }

 \vspace{-3mm}
	\caption{Real-world visual results on \textbf{\textit{dehazing}} and \textbf{\textit{deraining}}. Best viewed with zoom in. }
	\label{fig:visual_result}
 \vspace{-3mm}
\end{figure*}

\begin{figure*}[!htbp]
	\captionsetup{font=small}
	\scriptsize
	\centering
	
	\newcommand{\h}{0.105}
	\newcommand{\wa}{0.12}
	\newcommand{\wb}{0.16}
	\newcommand{\g}{-0.7mm}
   
 	\setlength\tabcolsep{1.8pt}
	\renewcommand{\arraystretch}{1}
	\resizebox{1.00\linewidth}{!} {
			\renewcommand{\h}{0.15}
			\newcommand{\w}{0.200}
				\begin{adjustbox}{valign=t}
					\begin{tabular}{ccccccc}
						\includegraphics[height=\h \textwidth, width=\w \textwidth]{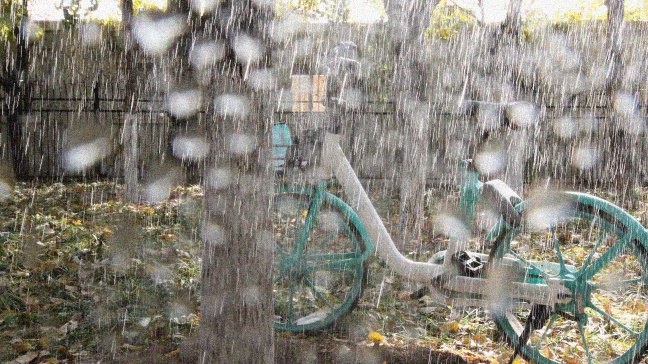} \hspace{\g} &
						\includegraphics[height=\h \textwidth, width=\w \textwidth]{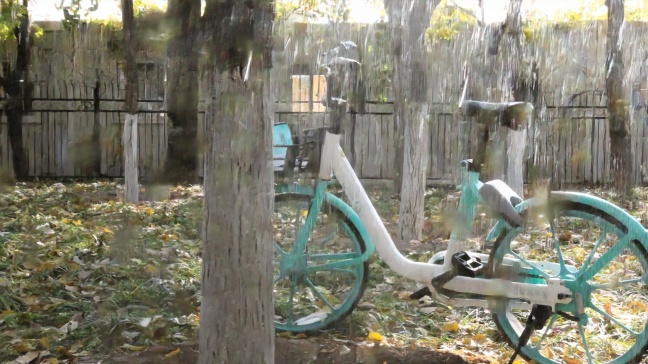} \hspace{\g} &
						\includegraphics[height=\h \textwidth, width=\w \textwidth]{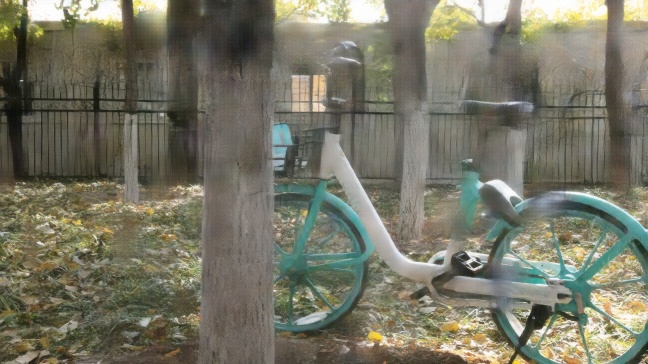} \hspace{\g} &
                        \includegraphics[height=\h \textwidth, width=\w \textwidth]{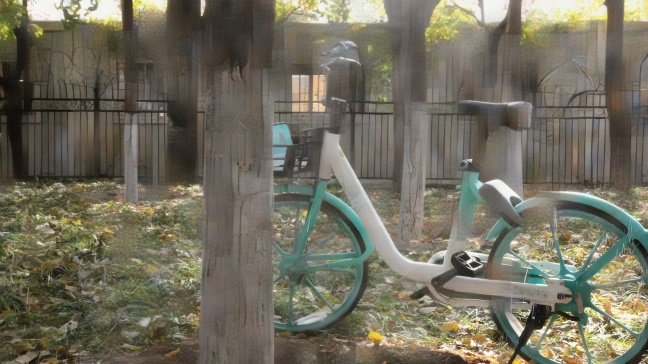} \hspace{\g} &
                        \includegraphics[height=\h \textwidth, width=\w \textwidth]{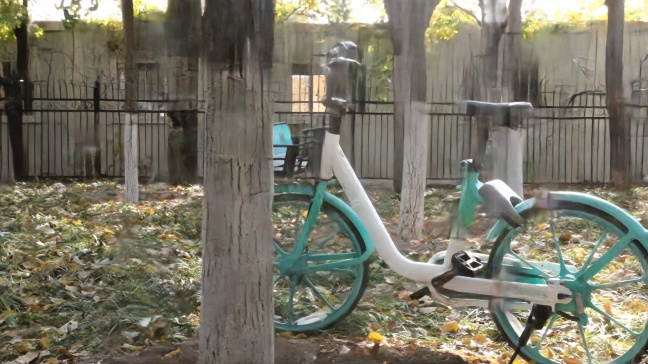} \hspace{\g} &
                        \includegraphics[height=\h \textwidth, width=\w \textwidth]{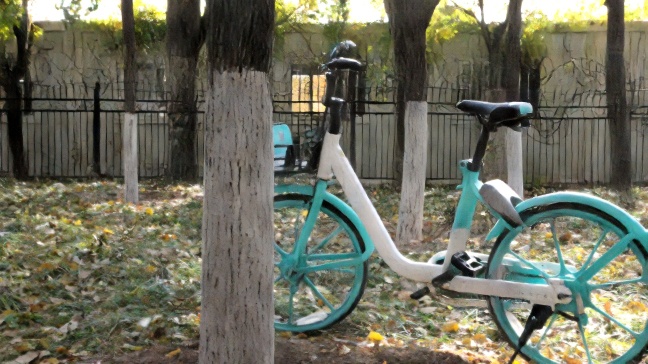} \hspace{\g} &
                        \includegraphics[height=\h \textwidth, width=\w \textwidth]{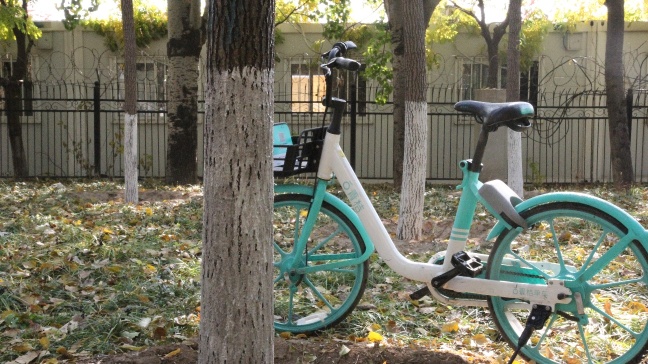}  
						\\
						R\&RD\&N Input / 13.01 dB \hspace{\g} &
                        AirNet~\cite{airnet} / 17.54 dB \hspace{\g} & 
                        TransWeather~\cite{valanarasu2022transweather} / 19.49 dB \hspace{\g} & 
                        WGWS-Net~\cite{zhu2023learning} / 19.62 dB \hspace{\g} & 
                        Restormer$_A$~\cite{nafnet} / 19.99 dB \hspace{\g} & 
                        \textbf{Ours} / 20.05 dB \hspace{\g} & 
                        GT / PSNR 
						\\
					\end{tabular}
				\end{adjustbox}
		 }

 	\setlength\tabcolsep{1.8pt}
	\renewcommand{\arraystretch}{1}
	\resizebox{1.00\linewidth}{!} {
			\renewcommand{\h}{0.15}
			\newcommand{\w}{0.200}
				\begin{adjustbox}{valign=t}
					\begin{tabular}{ccccccc}
						\includegraphics[height=\h \textwidth, width=\w \textwidth]{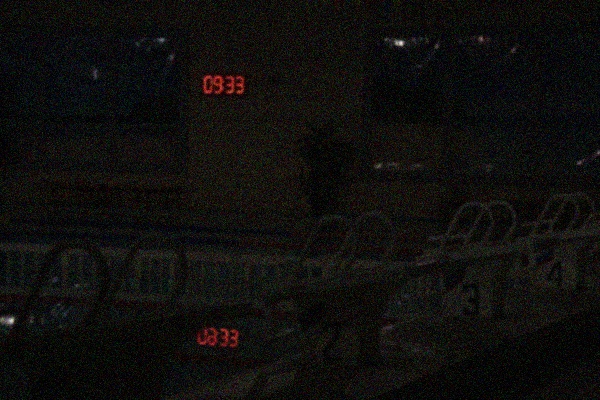} \hspace{\g} &
						\includegraphics[height=\h \textwidth, width=\w \textwidth]{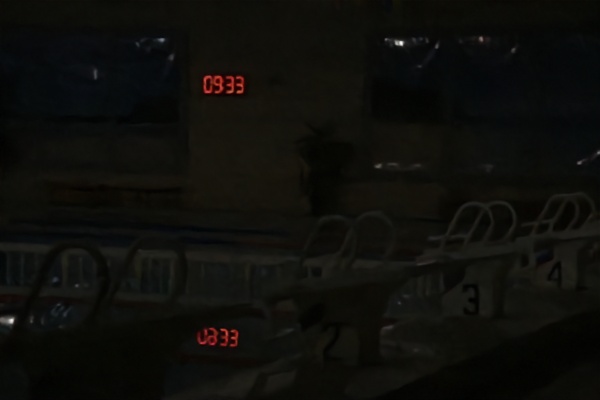} \hspace{\g} &
						\includegraphics[height=\h \textwidth, width=\w \textwidth]{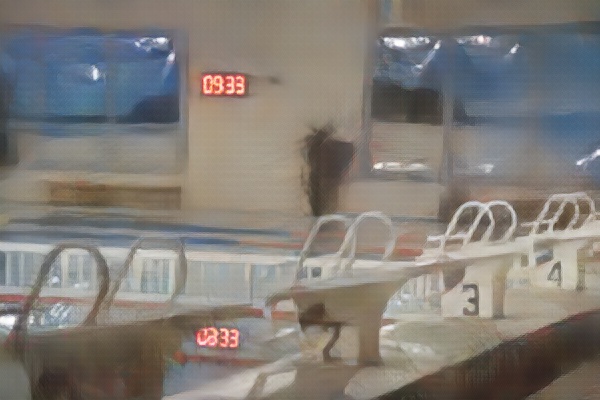} \hspace{\g} &
                        \includegraphics[height=\h \textwidth, width=\w \textwidth]{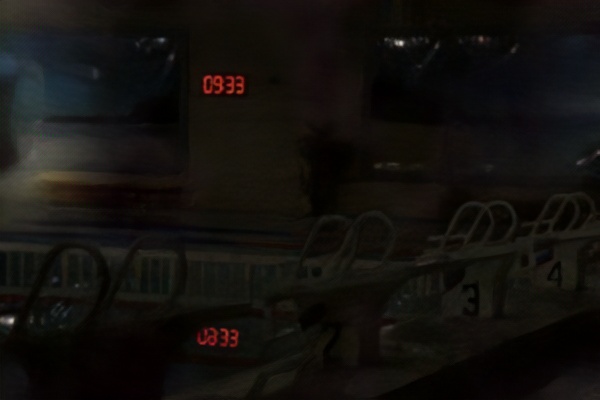} \hspace{\g} &
                        \includegraphics[height=\h \textwidth, width=\w \textwidth]{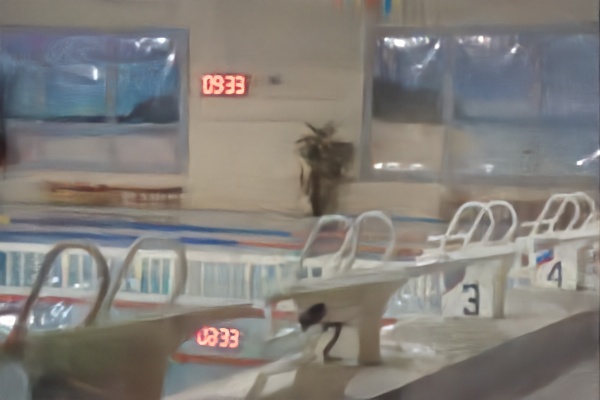} \hspace{\g} &
                        \includegraphics[height=\h \textwidth, width=\w \textwidth]{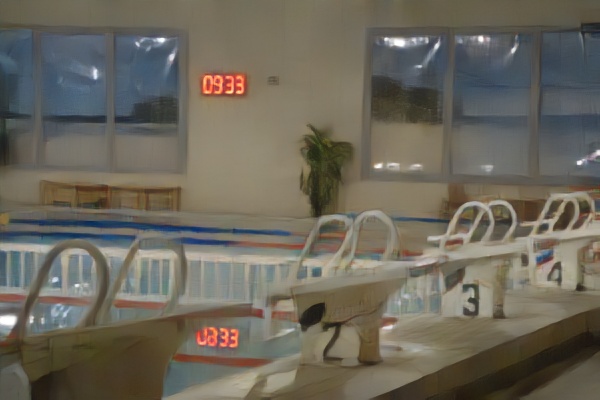} \hspace{\g} &
                        \includegraphics[height=\h \textwidth, width=\w \textwidth]{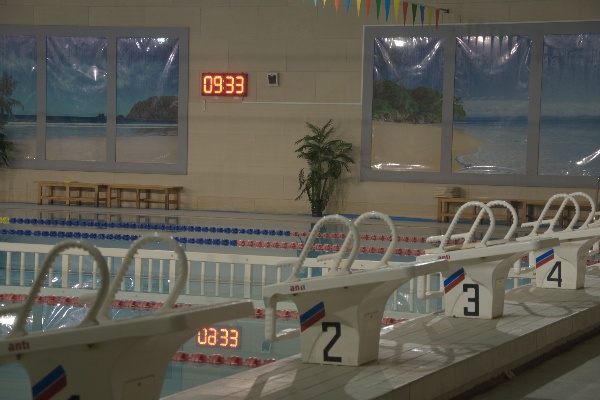}  
						\\
						LL\&N\&B Input / 10.12 dB \hspace{\g} &
                        AirNet~\cite{airnet} / 10.10 dB \hspace{\g} & 
                        TransWeather~\cite{valanarasu2022transweather} / 16.42 dB \hspace{\g} & 
                        WGWS-Net~\cite{zhu2023learning} / 10.36 dB \hspace{\g} & 
                        Restormer$_A$~\cite{nafnet} / 16.29 dB \hspace{\g} & 
                        \textbf{Ours} / 23.42 dB \hspace{\g} & 
                        GT / PSNR 
						\\
					\end{tabular}
				\end{adjustbox}
		 }
 
 \vspace{-3mm}
	\caption{Visual results on the MID6 benchmark (R: Rain; RD: RainDrop; N: Noise; LL: Low-Light; B: Blur). Our method can better handle these challenging cases where LQ images are affected by mixed degradations compared with other all-in-one methods. }
	\label{fig:visual_result_mid}
 \vspace{-5mm}
\end{figure*}

Table~\ref{tab:compare_9task} illustrates comprehensive performance comparisons with SOTA methods across 9 tasks. Our method consistently outperforms the compared all-in-one methods on all datasets. Notably, as an all-in-one approach, our method even achieves superior results compared to other task-specific methods in many tasks. Additionally, Table~\ref{tab:compare_mid} presents a comparison on the proposed MID6 benchmark, where our method demonstrates significant advantages in addressing challenging mixed-degraded images. Visual results of some intricate cases from MID6 are presented in Fig.~\ref{fig:visual_result_mid}, showcasing the enhanced effectiveness of our method in handling mixed degradations. Recognizing the significance of real-world IR challenges,  we present more results on real-world datasets in Table~\ref{tab:compare_real}. Fig.~\ref{fig:visual_result} offers visual comparisons across various tasks in real-world scenarios, showcasing the superiority of our method in addressing complex authentic degradations.

\begin{figure}[t]
\centering
\begin{subfigure}[b]{0.48\linewidth}
\centering
   \includegraphics[width=1\linewidth]{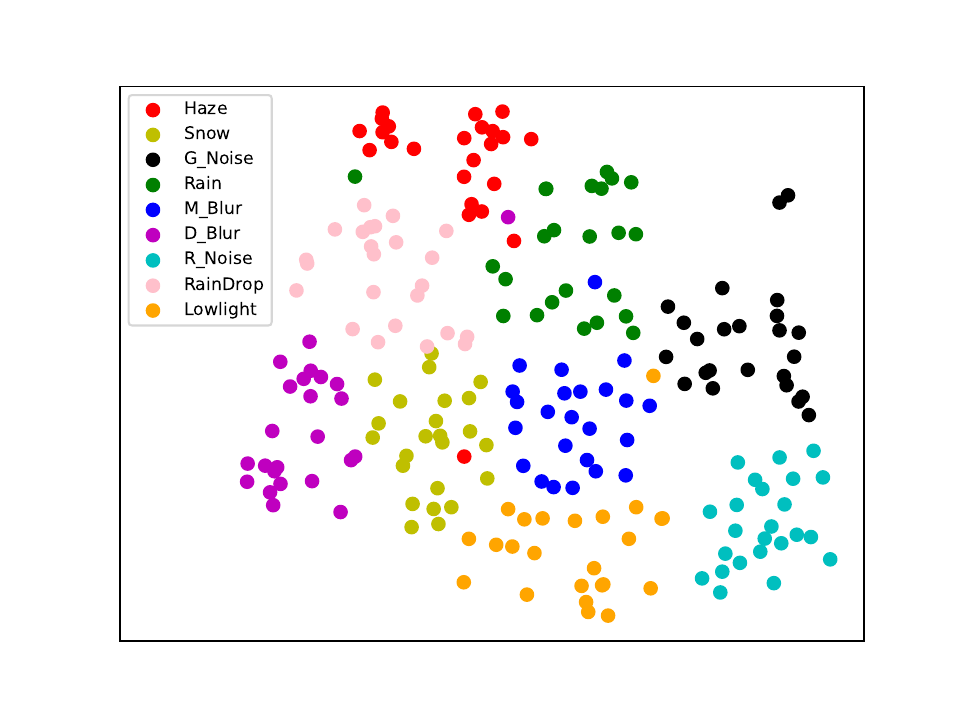}
   \captionsetup{font=tiny}
      \caption{t-SNE from visual prompt}
\end{subfigure}
\begin{subfigure}[b]{0.48\linewidth}
\centering
   \includegraphics[width=1\linewidth]{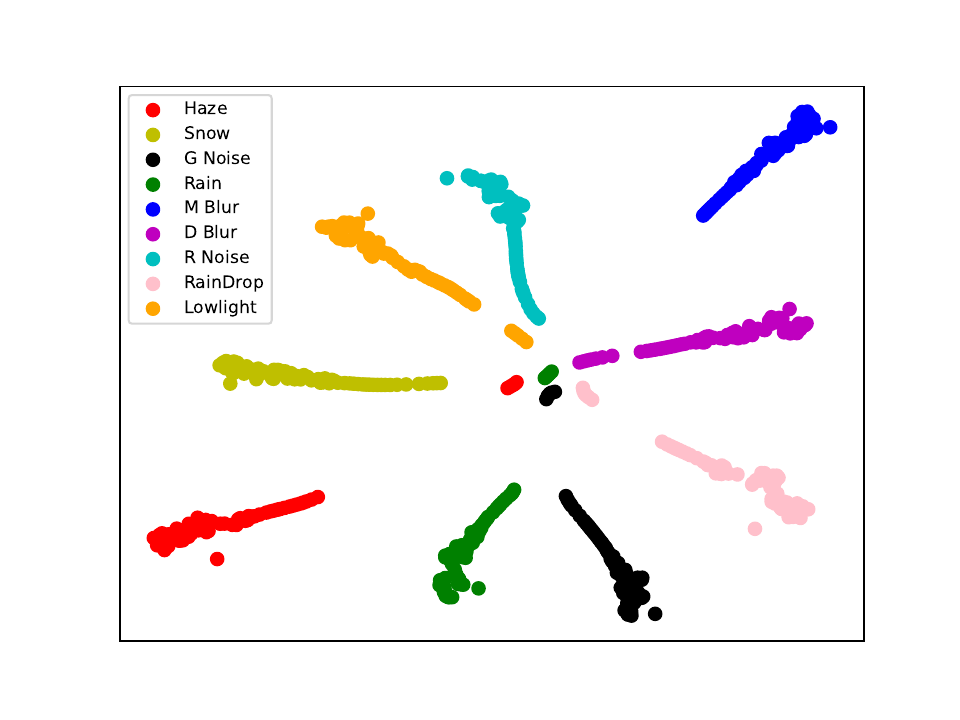}
   \captionsetup{font=tiny}
    \caption{t-SNE from CLIP textual embeddings}

\end{subfigure}
\vspace{-3mm}
\caption{t-SNE visualizations of visual prompt $V^1$ and CLIP textual embeddings $\mathcal{E}_{clip}^{txt}(T)$.}
\vspace{-3.5mm}
\label{fig:tsne}
\end{figure}

\begin{table}[thb]
\centering
\caption{Ablations of  MPerceiver. The metrics are reported on the average of deraining, dehazing and raindrop removal.}
\vspace{-3mm}
\label{tab:ablation}
\resizebox{0.46\textwidth}{!}{
\begin{tabular}{lccc}
\toprule
Method & PSNR $\uparrow$ & SSIM $\uparrow$ & LPIPS $\downarrow$ \\
\midrule
Baseline (Stable Diffusion) & 16.49 &  0.481 & 0.538 \\
\midrule
+Textual Branch & 19.19 &0.557 &0.447\\
+Textual Branch w/o TP Pool & 18.94 &0.553 &0.457 \\
\midrule
+Visual Branch & 25.05 &0.763 &0.213 \\
+Visual Branch w/o VP Pool & 24.94 &0.760 &0.216 \\
\midrule
+(Visual \& Textual) Branch & 25.31 &0.770 &0.199 \\
+(Visual \& Textual) Branch w/o TP Pool & 25.20& 0.768& 0.206 \\
+(Visual \& Textual) Branch w/o VP Pool & 25.13& 0.767& 0.204 \\
\midrule
+(Visual \& Textual) Branch + DRM (Full Model) & 29.17& 0.842& 0.162 \\
\bottomrule
\end{tabular}}
\vspace{-3mm}
\end{table}

\noindent\textbf{Zero-shot.} 
As shown in Tables~\ref{tab:zs_udc}\&\ref{tab:zs_uir}, we evaluate the performance of each all-in-one method on training-unseen tasks. MPerceiver outperforms compared methods in all metrics, demonstrating the generalizability of our approach.

\noindent\textbf{Few-shot.} 
As depicted in Tables~\ref{tab:compare_fs_jpeg}\&\ref{table:fs_mosaick}\&\ref{tab:compare_fs_moire}, we fine-tune the all-in-one methods with limited data to tailor them for new tasks. Notably, our method achieves comparable or superior results compared to task-specific methods trained with substantial amounts of data. This underscores the efficacy of MPerceiver, demonstrating that, post multitask pretraining, it has acquired general representations in low-level vision, allowing for cost-effective adaptation to new tasks.

\subsection{Ablation Study}
We perform ablation studies to examine the role of each component in MPerceiver. In Table~\ref{tab:ablation}, we initiate with SD and systematically incorporate or exclude the remaining modules of MPerceiver, including the visual branch, visual prompt (VP) pool, textual branch, textual prompt (TP) pool, and detail refinement module (DRM). The results exhibit a gradual improvement upon the addition of each component and a corresponding decline upon its removal, underscoring the effectiveness of each module.
Besides, Fig.~\ref{fig:tsne} visualizes the t-SNE statistics of visual prompt $V^1$ and CLIP textual embeddings $\mathcal{E}_{clip}^{txt}(T)$. It demonstrates that our multimodal prompt learning can effectively enable the network to distinguish different degradations.

\section{Conclusion}
This paper introduces MPerceiver, a multimodal prompt learning approach utilizing Stable Diffusion priors for enhanced adaptiveness, generalizability, and fidelity in all-in-one image restoration. The novel dual-branch module, comprising the cross-modal adapter and image restoration adapter, learns holistic and multiscale detail representations. The adaptability of textual and visual prompts is dynamically tuned based on degradation predictions, enabling effective adaptation to diverse unknown degradations. Additionally, a plug-in detail refinement module enhances restoration fidelity through direct encoder-to-decoder information transformation. Across 16 image restoration tasks, including all-in-one, zero-shot, and few-shot scenarios, MPerceiver demonstrates superior adaptiveness, generalizability, and fidelity.

\vspace{1mm}

\noindent\textbf{Acknowledgements:} This research is partially funded by Youth Innovation Promotion Association CAS (Grant No. 2022132), Beijing Nova Program (20230484276), National Natural Science Foundation of China (Grant No. U21B2045, U20A20223) and CAAI Huawei MindSpore Open Fund.

{    \small
    \bibliographystyle{ieeenat_fullname}
    \bibliography{main}
}


\end{document}